\documentclass{article}

\usepackage[english]{babel}

\usepackage[letterpaper,top=2cm,bottom=2cm,left=3cm,right=3cm,marginparwidth=1.75cm]{geometry}
\usepackage[breaklinks=true]{hyperref}
\usepackage[anythingbreaks]{breakurl}
\usepackage{etoolbox}

\newtheorem{remark}{Remark}
{\bf}{}
\BeforeBeginEnvironment{table}{}
\usepackage{algorithm}
\usepackage[noEnd=true,indLines=true]{algpseudocodex}

\algrenewcommand\algorithmicrequire{\textbf{Input:}}
\algrenewcommand\algorithmicensure{\textbf{Output:}}
\algnewcommand\algorithmicparameters{\textbf{Parameters:}}

\makeatletter
\algnewcommand\Parameters{%
    \algpx@endCodeCommand
    \algpx@drawInItem{\algorithmicparameters}%
}
\makeatother

\usepackage{color}
\usepackage{amsfonts}
\usepackage{amssymb}
\usepackage{amsmath}
\usepackage[utf8]{inputenc}
\usepackage[T1]{fontenc}
\usepackage{graphicx}
\usepackage{booktabs}
\usepackage{subfigure}
\usepackage{lscape}
\usepackage{mdframed} % This one is new!!!
\usepackage{adjustbox}
\usepackage{todonotes} 
\usepackage{float}
\usepackage{enumitem} 
\usepackage{setspace}
\usepackage{comment}
\usepackage{tikz}
\usetikzlibrary{shapes,arrows}
\usetikzlibrary{calc}

\newcommand{\x}{{\boldsymbol{x}}}            % vector x bolded
\newcommand{\y}{{\boldsymbol{y}}}            % vector y bolded
\newcommand{\ba}{{\boldsymbol{a}}}           % vector a bolded
\newcommand{\Jobj}{{f}}              % Objective function
\newcommand{\dimopti}{{N}}
\newcommand{\numpart}{{M}} % klusterointi osassa pieni m

\newcommand{\fk}{f^k}              % clustering function with index k
\newcommand{\bxi}{\boldsymbol{\xi}}          % vector \xi bolded
\newcommand{\batch}{\mathcal{B}}          % vector 0 bolded
\newcommand{\sample}{\mathcal{S}}          % vector 0 bolded
\newcommand{\isic}{{\sc ISIC }}
\newcommand{\isicns}{{\sc ISIC}} % no space after the name
\newcommand{\gspa}{{\sc GSPA }}
\newcommand{\gspans}{{\sc GSPA}} % no space after the name
\newcommand{\islmbm}{{\sc SILMB }}
\newcommand{\islmbmns}{{\sc SILMB}} % no space after the name
\newcommand{\ilmbm}{{\sc InexactLMBM }}
\newcommand{\ilmbmns}{{\sc InexactLMBM}} % no space after the name
\newcommand{\LMBM}{{\sc LMBM }}
\newcommand{\LMBMNS}{{\sc LMBM}}

\newcommand{\slmbmns}{{\sc SLMBA}}
\newcommand{\slmbm}{{\sc SLMBA }}
\newcommand{\msmgkm}{{\sc MS-MGKM }}

\newcommand{\mbkm}{{\sc MiniBatchKMeans }}

\newcommand{\dcclust}{{\sc DC-Clust }}

\newcommand{\sclust}{{\sc BigClust }}
\newcommand{\sclustns}{{\sc BigClust}}
\newcommand{\bigClust}{{\sc BigClust }}
\newcommand{\bigClustns}{{\sc BigClust}}
\newcommand{\lmbclust}{{\sc LMB-Clust }}
\newcommand{\lmbclustns}{{\sc LMB-Clust}}
\newcommand{\bigmeans}{{\sc Big-Means }}
\newcommand{\bigmeansns}{{\sc Big-Means}}
\newcommand{\hgmeans}{{\sc HG-Means }}

\newcommand{\drsmeans}{{\sc DRS-Means }}

\newcommand{\R}{\mathbb{R}}

\DeclareMathOperator*{\argmin}{arg\,min}
\DeclareMathOperator*{\argmax}{arg\,max}

\title{ISIC --- Incremental Clustering Software Based on Stochastic Inexact Nonsmooth Optimization
       }
\author{%
Napsu Karmitsa\textsuperscript{1}\thanks{\texttt{napsu@karmitsa.fi}},
Jenni Lampainen\textsuperscript{2},
Kaisa Joki\textsuperscript{2},\\
Marko M. Mäkelä\textsuperscript{2},
Antti Airola\textsuperscript{1},
Tapio Pahikkala\textsuperscript{1}\\[1ex]
{\small \textsuperscript{1}Department of Computing,
University of Turku, FI-20014 Turku, Finland}\\
{\small \textsuperscript{2}Department of Mathematics and Statistics,
University of Turku, FI-20014 Turku, Finland}
}
\begin{document}
\maketitle
\begin{abstract}
Clustering is a fundamental task in data mining and machine learning, but solving minimum sum-of-squares clustering problems is computationally demanding for very large and high-dimensional datasets. In this paper, we present \isicns, an incremental clustering algorithm based on stochastic inexact nonsmooth optimization. 
\isic builds on and modifies the existing incremental clustering framework by introducing two new components. First, it introduces a new greedy starting-point algorithm that directly constructs a promising starting center for each new cluster, avoiding the costly multi-start procedure of earlier incremental approaches and eliminating the need to solve separate auxiliary clustering problems. Second, it employs the novel stochastic inexact limited-memory bundle method (\islmbmns) as its optimization engine. By explicitly accommodating inexact function and subgradient information, \islmbm enables \isic to better account for inaccuracies introduced by stochastic computations.
The resulting incremental approach is designed to produce cost-efficiently high-quality, near-global solutions and provides solutions not only to the final $k$-partition problem but also to all intermediate partitioning problems.
This is particularly useful when the optimal number of clusters is
not known in advance.
We evaluate \isic on several real-world datasets and compare its performance with clustering methods designed for large and big data. The numerical results demonstrate that \isic efficiently produces high-quality clustering solutions while reducing the computational effort required by other incremental clustering approaches.
\smallskip

\end{abstract}

\section{Introduction} 
Clustering is a central task in data mining and machine learning, with applications in areas such as medicine and bioinformatics \cite{PSO1,bioinf1,bioinf2}, cybersecurity \cite{cybersec1,cybersec2}, text mining \cite{text_mining1}, image processing \cite{9151332}, and economics, finance, and marketing \cite{reutterer2021cluster}. 
It aims to partition a dataset into clusters so that points within the same cluster are more similar to each other than to points in different clusters. 

In this paper, we consider the \emph{hard clustering problem}, where each data point is assigned to exactly one cluster. Similarity between data points can be measured in several ways, but the squared Euclidean distance adapted in this paper is among the most widely used options. These choices lead to the \emph{minimum sum-of-squares clustering (MSSC) problem}, 
which minimizes the total squared distance from each point to its assigned cluster center (see, e.g., \cite{BagKarTah:2025}). 
Further, we consider an \emph{incremental clustering} approach, which starts with one cluster center representing the entire dataset and adds one new cluster center at a time until the prescribed maximum number of clusters $k$ is reached. At each incremental step, the solution obtained for the current number of clusters is used to construct a favorable starting configuration for the next clustering problem, helping to obtain high-quality clustering solutions. As an additional benefit, the approach produces solutions to all $l$-partition problems, $l=1,\ldots,k$, within a single incremental procedure. When the appropriate number of clusters is not known in advance, these intermediate solutions can also be used to select a suitable number of clusters, for example, by applying cluster validity indices (see, e.g., \cite{BagKarTah:2025,BagAliSUl2023,BagUgoWeb:2011}).

The MSSC problem is an NP-hard global optimization problem \cite{MSSC_NPhard}. An exact algorithm based on semidefinite programming (SDP) relaxations and branch-and-bound is presented in \cite{sos-sdp}; 
however, for $m$ data points this algorithm involves an $m\times m$ SDP matrix variable and the computational cost of the branch-and-bound procedure grows rapidly with $m$. 
Thus, this type of approach is computationally prohibitive for large datasets. 
In their turn, heuristic approaches such as the $k$-means  and its numerous variants
are widely used for the MSSC problem
(see, e.g., 
\cite{BagUgoWeb:2011,DavVas:2007,Ism2018,MusMlaJarMus2023,KozMus2024,beyondkmeans2024,VolTolWeb:2013}).
 Although computationally efficient, these methods are often sensitive to the choice of initial cluster centers and may converge to poor local solutions, especially for larger datasets.

At the same time, the size and dimensionality of modern datasets continue to increase as advances in computing and data storage make it possible to collect and process ever larger amounts of data. This creates an increasing need for scalable clustering methods that can maintain high solution quality while keeping the computational cost manageable when clustering very large and big\footnote{We use the definition of big data given in \cite{MusMlaJarMus2023}: big data is a dataset of such a massive volume that its processing causes significant technical difficulties or becomes impossible using traditional methods and average computing resources.} datasets.
To address this need,
various optimization-based approaches have been
developed for the MSSC problem, such as 
\vspace{-1.2ex}
\begin{itemize}
\item nonsmooth optimization-based clustering methods \cite{BagKarTah:2025,BagAliSUl2023,KarBagTah:2018,bigClust:2025,LamJokKar:2026} %; \vspace{-1.2ex}
including
%\item 
approaches based on difference-of-convex representations of the clustering function \cite{BagKarTah:2025,BagTahUgo:2016,LeTLeHPha:2014,KhaAstDAlGau:2016};
%\cite{BagKarTah:2025,BagTahUgo:2016,LeTLeHPha:2014,KarBagTah:2017,KhaAstDAlGau:2016}; 
\vspace{-1.2ex}

\item hyperbolic smoothing techniques \cite{BagOrdOztXav:2015,XavXav2020}; and \vspace{-1.2ex}

\item metaheuristics, including simulated annealing \cite{SeiBagBorPic2019,SelAlS:1991}, tabu search \cite{Al-Sultan1995,tabu1}, particle swarm optimization \cite{PSO1,PSO2}, and evolutionary algorithms \cite{GriVid2019,ManSch2021}. \vspace{-1.2ex}
\end{itemize}
\noindent
Among these approaches, nonsmooth optimization (NSO) is particularly attractive for large-scale MSSC problems. NSO concerns the minimization or maximization of functions whose gradients may be discontinuous \cite{BagKarMak:2014}.
A key advantage of the NSO formulation of MSSC is that the number of  optimization  
variables $N$ depends on the number of clusters $k$ and features $n$, but not directly on the number of data points $m$. By contrast, standard mixed-integer programming formulations typically require $mk$ binary 
and $nk$ continuous variables \cite{BagKarTah:2025}, which scales poorly with $m$.

Consequently, several NSO-based incremental clustering methods have been developed for large datasets. 
In \cite{KarBagTah:2018}, the incremental clustering approach originally proposed by Ordin and Bagirov \cite{OrdBag:2015} was combined with the limited-memory bundle method (\LMBMNS) \cite{HaaMieMak:2004,HaaMieMak:2007}, resulting in the deterministic \lmbclust algorithm. More recently, \bigClust \cite{bigClust:2025} extended this framework to very large and big datasets by introducing stochasticity into both the incremental procedure and the underlying NSO process. Its stochastic limited-memory bundle algorithm (\slmbmns) uses mini-batch information to reduce the computational cost of solving the optimization problems arising within the incremental clustering framework.

The underlying incremental strategy \cite{OrdBag:2015} --- 
later used in several clustering algorithms (see, e.g., \cite{BagAliSUl2023,KarBagTah:2018,bigClust:2025,BagTahUgo:2016,BagOrdOztXav:2015,KarBagTah:2017}, and \cite{BagKarTah:2025} for an overview) --- aims to reduce the sensitivity of local optimization methods to the initial cluster centers and steer the optimization toward high-quality, near-global solutions. At each incremental step, a set of candidate starting points for the new cluster center is first extracted from the data. Keeping the existing cluster centers fixed, an auxiliary clustering problem is solved separately from each of these points to obtain a set of improved candidate centers for the new cluster. Each improved candidate is then appended to the existing cluster centers, producing a set of starting configurations for the corresponding main clustering problem with one additional cluster. This problem is solved from each of these configurations, and the best resulting solution is used as the basis for the next incremental step. Although this multi-start strategy has been shown to produce high-quality solutions, repeatedly solving both the auxiliary and main clustering problems from multiple starting points can become computationally expensive for very large datasets.

The method proposed in this paper, \isicns, builds on the stochastic incremental clustering framework of \bigClust \cite{bigClust:2025}, but modifies it substantially in two main
respects.
First, it introduces the greedy starting-point algorithm, \gspans,
which evaluates the improvement potential of candidate points, selects the candidate providing the largest immediate improvement, and directly constructs a promising new center based on this candidate. 
Thus, \isic replaces the computationally expensive multi-start procedure described above, and eliminates the need to solve separate auxiliary clustering problems.
Second, \isic employs \islmbmns, a novel stochastic variant of \ilmbm \cite{inexactLMBM2026}, instead of the stochastic limited-memory bundle method used in \bigClustns. \islmbm serves as the optimization engine throughout the incremental clustering procedure. \ilmbm is a limited-memory bundle method designed for large-scale NSO with inexact function and subgradient information. By explicitly accommodating such inexact information, \islmbm allows \isic to better account for the inaccuracies introduced by stochastic computations. 
Together, these modifications retain the idea of identifying promising new cluster centers incrementally while substantially reducing the computational effort required at each incremental step. 
The overall structure of \isic is given in Figure~\ref{flowchart}.\medskip

%-----------FLOWCHART ALKAA-----------

% Define block styles

\tikzstyle{decision1} = [diamond, draw, fill=blue!10, 
text width=5em, text badly centered, inner sep=0pt]

\tikzstyle{alku} = [rectangle, draw, fill=blue!10,
    text width=10em, text centered, minimum height=3em,line width=1.0mm]
    
\tikzstyle{block1} = [rectangle, draw, fill=blue!10,
    text width=10em, text centered, minimum height=3em]
    
\tikzstyle{block2} = [rectangle, draw, fill=blue!10,
    text width=13em, text centered, minimum height=4.6em]

\tikzstyle{end} = [rectangle, draw, fill=blue!10, line width=0.7mm, text width=4.4em, text centered, minimum height=2.5em]
    
\tikzstyle{line} = [draw, line width=0.45mm, color=black!100, -latex']

\begin{figure}[h!]
\centering
\resizebox{0.85\linewidth}{!}{
\begin{tikzpicture}[node distance = 1.5cm, auto]

%------- Place nodes--------

    \node [alku] (vasen1) {\textbf{\,Initialization}\\ Start with one cluster};

    \node [block1, below of=vasen1, node distance=2.5cm] (vasen2) {Choose/update the sample set};

    \node [block2, right of=vasen2, aspect=2, node distance=5.4cm] (keski1) {Apply \textbf{GSPA} to find\\a \textbf{starting point} for\\the clustering problem};
    
    \node [block2, right of=keski1, node distance=6.0cm] (oikea1) {Apply \textbf{ISLMB}\\to solve the\\\textbf{clustering problem}};
    
    \node [decision1, below of=keski1, node distance=3.0cm] (keski2) {Enough clusters?};

    \node [block1, below of=vasen2, aspect=2, node distance=3.0cm] (vasen3) {Increase the number\\of clusters by one};

    \node [end, below of=oikea1, node distance=3.0cm] (lopetus) {\textbf{STOP}};

% --- NUOLET ---
\path [line] (vasen1) -- (vasen2);
\path [line] (vasen2) -- (keski1);
\path [line] (keski1) -- (oikea1);
\path [line] (oikea1.south west) -- (keski2.north east);
\path [line] (keski2) -- (vasen3) node[midway, above]{No};
\path [line] (vasen3) -- (vasen2);
\path [line] (keski2) -- (lopetus) node[midway, above]{Yes};
    
\end{tikzpicture}
}
\caption{Flowchart of ISIC for MSSC problems.}
\label{flowchart}
\end{figure}
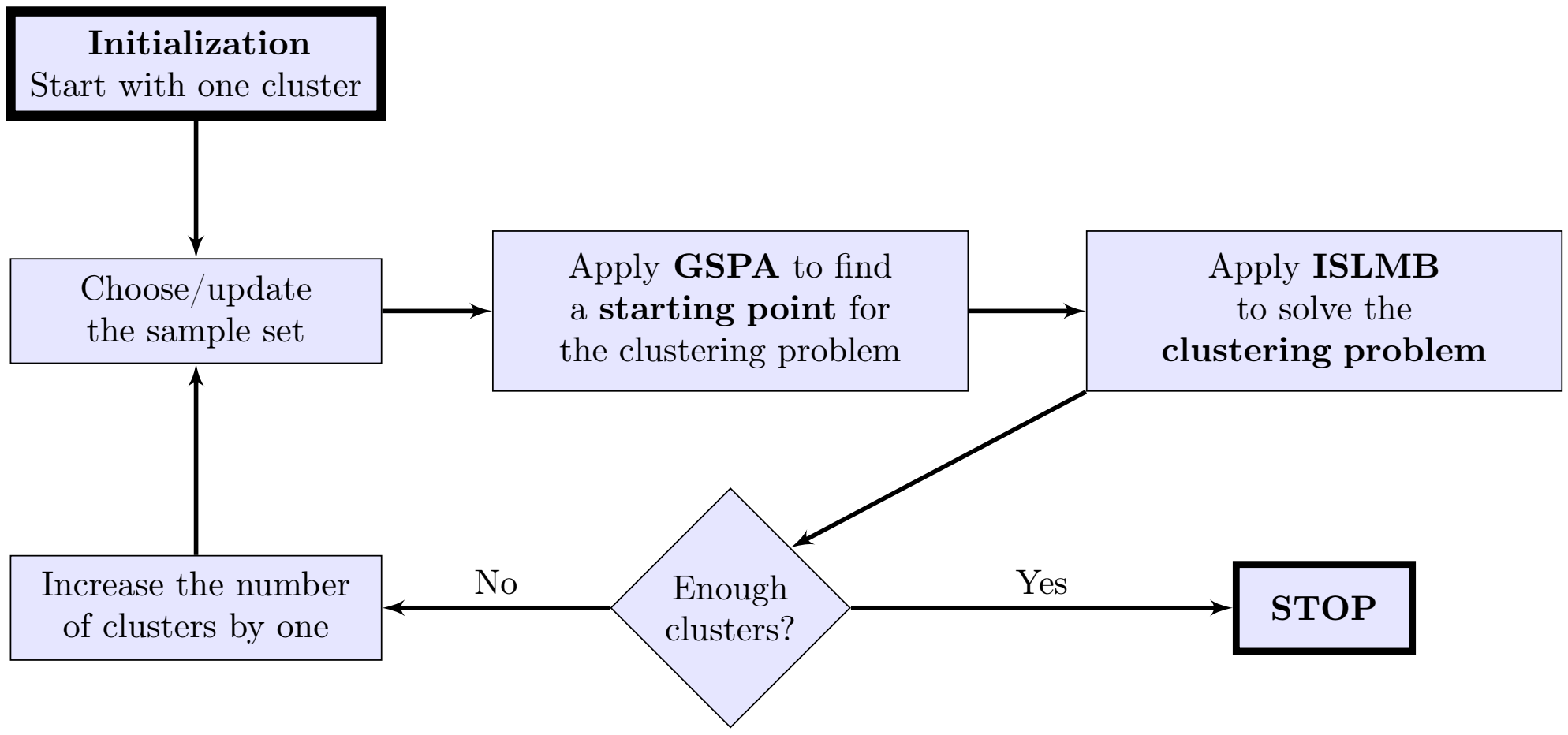

%-----------FLOWCHART LOPPUU----------

The main contributions of this paper are as follows:
\begin{enumerate}
\item We introduce \islmbmns, a stochastic variant of the inexact limited-memory bundle method \ilmbmns, for solving large-scale nonsmooth finite-sum optimization problems with stochastic and inexact information. As an important
example of this problem class, we consider the MSSC problem.

\item We introduce \gspans, a greedy starting-point algorithm that simplifies
the incremental clustering procedure by directly constructing a promising starting
center for each new cluster.

\item We propose \isicns, an incremental clustering algorithm for very large and big MSSC problems that employs \islmbm as its optimization engine and \gspa to efficiently generate starting points for the underlying clustering problems. 

\item We conduct an extensive numerical evaluation of \isic on very large and big datasets. The main comparisons are made with \lmbclust
\cite{KarBagTah:2018} and \bigClust \cite{bigClust:2025}, the closest
predecessors of \isic within the NSO-based incremental clustering framework, and with
\bigmeans \cite{MusMlaJarMus2023}, which provides a scalable $k$-means-based
benchmark. Comparisons with several additional clustering methods are
provided in Supplementary Material 2.

\item We provide an open-source implementation of \isic at \url{https://github.com/napsu/ISIC}.
\end{enumerate}

The remainder of this paper is organized as follows. Section~\ref{sec_clus} presents the clustering formulation used in \isicns. The complete \isic algorithm is introduced in Section~\ref{sec_isic}, followed by \gspa in Section~\ref{sec_gspa}, and \islmbm in Section~\ref{sec_islmb}.
Numerical experiments are reported in Section~\ref{sec_num}, and, finally, Section~\ref{sec_conclusion} concludes the paper.

\section{Minimum sum-of-squares clustering} \label{sec_clus}

In this section, we present the {\em NSO formulation of the MSSC problem} and show how it can be represented as a finite-sum optimization problem. For more
in-depth information on NSO-based clustering, we refer to \cite{BagKarTah:2025}.

\subsection{Cluster analysis} 
Let a dataset \[ A=\{\ba_1,\ldots,\ba_m\}, \qquad \text{where}\,\,\, \ba_i\in\R^n,\,\,\, i=1,\ldots,m, \] contain $m$ data points with $n$ features. In the {\em unconstrained hard clustering problem}, the aim is to partition $A$ into $k$ nonempty, pairwise disjoint clusters $A^1,\ldots,A^k$ such that 
\[ A=\bigcup_{j=1}^k A^j. \] 
%$A=\bigcup_{j=1}^k A^j.$ 
Note that each data point $\ba_i$ belongs to exactly one cluster, and each cluster $A^j$ is represented by a center $\x_j\in\R^n$.

In this paper, {\em similarity} is measured using the squared Euclidean distance \[ d(\x,\ba)=\|\x-\ba\|^2, \qquad \text{where }\,\x,\ba\in\R^n. \] 
%The corresponding clustering problem is called the {\em minimum sum-of-squares clustering (MSSC) problem}.
%
Using this similarity measure, the MSSC problem --- the {\em $k$-clustering problem} --- can be formulated as the NSO problem
\cite{BagKarTah:2025} 
\begin{align}
  \label{nonsmooth opt1}
  \begin{cases}
    \text{minimize}\quad  & \displaystyle \fk(\x_1,\ldots,\x_k)=\frac{1}{m}\sum_{i=1}^m \min_{j=1,\ldots,k} d(\x_j,\ba_i) \\
    \text{subject to} &(\x_1,\ldots,\x_k) \in \R^{nk},
  \end{cases}
\end{align}
whose solution is denoted by $(\x_1^k,\ldots,\x_k^k)$. In what follows, we may omit the superscript $k$ when there is
no risk of ambiguity.
The function $\fk$ is called the {\em $k$-cluster function}. It is locally Lipschitz continuous (LLC), nonconvex, and nonsmooth for $k>1$ \cite{BagKarTah:2025,BagKarMak:2014}; these and other basic concepts of NSO used in this paper are summarized in Supplementary Material 1. %~\ref{appendix}.

\subsection{Nonsmooth finite-sum formulation} \label{sec_finitesum}

% UUSI ALKAA %
Many objectives in machine learning and data analysis can be presented in the form
\begin{align}
    \label{sum-function}
    \Jobj(\x)=\frac{1}{\numpart} \sum_{i=1}^\numpart \Jobj_i(\x),
\end{align}
where $\x \in \R^\dimopti$ and the component functions $f_i$, $i\!=1,\ldots,\numpart$, are LLC and not necessarily smooth. This finite-sum formulation is suitable for stochastic optimization, where only a subset of the components are used at a time.
For a randomly selected index set 
$\batch\subseteq\{1,\ldots,M\}$ with $ |\batch|=m_\batch\leq M,$
we define the corresponding {\em batch function} by
$$
\Jobj_{\batch}(\x)=\frac{1}{ m_\batch} \sum_{i \in \batch} \Jobj_i(\x).
$$
Its subdifferential $\partial\Jobj_{\batch}(\x)$ is referred to as the {\em stochastic subdifferential}, and any $ \bxi_{\batch}\in\partial\Jobj_{\batch}(\x) $ is called a {\em stochastic subgradient}. Here, $\partial\Jobj(\x)$
denotes the {\em Clarke subdifferential} of function $f$ at $\x$ \cite{BagKarMak:2014,Cla:1983} (see Supplementary Material 1).

The NSO formulation of the MSSC problem fits directly into this framework, with one component function $f_i$ associated with each data point $\ba_i$. For the $k$-clustering problem \eqref{nonsmooth opt1}, these components are
\[
\Jobj_i^k(\x_1,\ldots,\x_k)
=
\min_{j=1,\ldots,k} d(\x_j,\ba_i), \qquad \text{for each }i=1,\ldots,m,
\]
so that $M=m$ and $N=nk$. Consequently, stochastic optimization can be applied to this problem by evaluating the corresponding objective functions and subgradients on batches of data rather than on the complete dataset. This is the finite-sum structure exploited by \islmbm in Section~\ref{sec_islmb}.

% UUSI LOPPUU %
\begin{comment}  
$$
\Jobj_i^k(\x_1,\ldots,\x_k)
$$
\end{comment}

\section{ISIC --- Incremental clustering software} \label{sec_isic}

Next, we introduce \isicns, an incremental clustering algorithm designed for
solving MSSC problems in very large and big datasets. \isic builds on the
stochastic incremental framework of \bigClust \cite{bigClust:2025}, but
modifies the incremental procedure by eliminating the separate auxiliary
clustering problems and the multi-start procedure used in earlier approaches. 
This is achieved through the greedy starting-point procedure \gspa
(Algorithm~\ref{alg_find_sp_greedy}), which directly constructs a new starting center for the next clustering problem. Thus, at each incremental step, whereas earlier methods solve both the
auxiliary and main clustering problems multiple times from different
starting points, \isic solves the clustering problem only once from a single starting configuration obtained using 
\gspans.
In addition, \isic introduces the stochastic
inexact limited-memory bundle method \islmbm
(Algorithm~\ref{alg_SILMBM}) as its optimization engine.

Let $k$ denote the prescribed maximum number of clusters and
$l \in \{ 2,\ldots,k\}$ the number of clusters at the current incremental step.
In \isicns, the transition from $l-1$ to $l$ clusters
consists of two principal stages.\footnote{Note that solving the $1$-clustering problem is a trivial task.} 
First, \gspa determines a starting center $\bar{\y}\in\R^n$ for the new
cluster based on a candidate point selected to yield the largest estimated
decrease in the clustering objective.
This center is appended to the current $(l-1)$-clustering solution, yielding
the initial point $(\x_1^{l-1},\ldots,\x_{l-1}^{l-1},\bar{\y})\in\R^{nl}$ %$(\x_1,\ldots,\x_{l-1},\bar{\y})\in\R^{nl}$ 
for the
$l$-clustering problem.
Second, this initial configuration is refined by minimizing the $l$-cluster function using \islmbmns. The resulting solution $(\x^l_1,\ldots, \x_l^l)$ is then
used in the next incremental step. This procedure is repeated until the
prescribed maximum number of clusters is reached, yielding solutions not only
to the final $k$-clustering problem but also to all intermediate
$l$-clustering problems, $l=1,\ldots,k-1$. These solutions can then be
compared using suitable cluster-validity criteria, such as the
Davies--Bouldin or Dunn index \cite{DavBou:1979,Dun:1974}, to support the selection of an appropriate
number of clusters~\cite{BagKarTah:2025}.

Similarly to \bigClust \cite{bigClust:2025}, \isic employs a doubly stochastic approach. At each incremental step, a uniform random sample $A_\sample \subseteq A$ of size $m_\sample \leq m$ can be selected and used in place of the full dataset $A$. In addition, the optimization problems arising within the incremental procedure are solved by \islmbm using randomly selected batches of size $m_\batch \leq m_\sample$. Thus, stochasticity is introduced both at the level of the incremental clustering procedure through the sample $A_\sample$ and within the optimization process through the batches used by \islmbmns.

Algorithm~\ref{algorithm_isic} summarizes the complete \isic procedure. 
{ %\color{red}
Here, \(\textsc{UniformSample}(X,s)\) returns a uniformly random subset of \(s\) distinct elements from the finite set \(X\), sampled without replacement.}

\begin{algorithm*} % ISIC
\caption{\isic}
\label{algorithm_isic}
{\small
\begin{algorithmic}[1]
\Require Dataset $A$; maximum number of clusters $k \geq 1$;
 sample size $ m_\sample \leq m$;
 batch size $ m_\batch \leq  m_\sample$; % n_batch = max(10.0 * SQRT(m_\sample),1000.0) 
 maximum number of batches $T_\batch \geq 1$. % maxbatch= MAX(20, 0.15 * SQRT(m_\sample))

\Ensure A solution  $(\x^l_1,\ldots,\x^l_{l}) \in \R^{nl}$ 
to the $l$-clustering problem $\min_{(\x_1,\ldots,\x_l)\in\R^{nl}}f^l(\x_1,\ldots,\x_l)$ for every $l=1,\ldots,k$.

\State $\mathcal{I}_{\sample}\leftarrow\Call{UniformSample}{\{1,\ldots,m\},m_S}$; \;\;   $ A_S\leftarrow\{\ba_i\mid i\in \mathcal{I}_{\sample}\}$; \;\; $l \leftarrow 1$
\Comment{Initialization}

\State $\x^1_1 \leftarrow \frac{1}{m_\sample} \sum_{i \in \mathcal{I}_{\sample}} \ba_i.$
\Comment{Solution of the 1-clustering problem on $A_{\sample}$}
\For{$l=2,\ldots,k$}
\State $\mathcal{I}_{\sample}\leftarrow\Call{UniformSample}{\{1,\ldots,m\},m_S}$; \;\;   $ A_S\leftarrow\{\ba_i\mid i\in \mathcal{I}_{\sample}\}$;
    \Comment{Update sample}
    \State $\bar{\y} \leftarrow \Call{GSPA}{{A_{\sample},(\x^{l-1}_1,\ldots,\x^{l-1}_{l-1}),l}}$
    \Comment{Starting center for new cluster}
    \State $(\x^l_1,\ldots,\x^l_{l}) \leftarrow \Call{SILMB}{(\x^{l-1}_1,\ldots,\x^{l-1}_{l-1},\bar{\y}),f^l_{A_\sample},m_\batch,T_{\batch}}, \; \text{where } f^l_{A_{\sample}}(\cdot)=\frac{1}{m_\sample}\sum_{i \in \mathcal{I}_{\sample}} f^l_i(\cdot)$ \\
    \Comment{Solve $l$-clustering problem \eqref{nonsmooth opt1} on $A_\sample$}
\EndFor
\State \Return $(\x^l_1,\ldots,\x^l_{l}) \in \R^{nl}$ for all $l=1,\ldots,k$
\end{algorithmic}}
\end{algorithm*}

\begin{remark}[Parameter choices]\label{rem:isic_parameters}
When computationally feasible, we recommend using $A_\sample=A$, since
restricting the incremental procedure to a smaller sample may make the
clustering more sensitive to variations in the sampled data. When
$A_\sample=A$, the incremental procedure operates on the complete dataset,
while the optimization process remains stochastic through the randomly
selected batches used by \islmbmns.

In our numerical experiments, the prescribed maximum number of clusters is
$k=25$. The remaining parameter choices described below were determined based
on preliminary experiments with \isicns.  
For the sample size, we consider both the complete dataset,
$m_\sample=m$, and subsampled variants with
\[
m_\sample=\max\left\{\left\lfloor\frac{m}{\#}\right\rfloor,\,10\,000\right\},
\qquad
\#\in\{10,20,50\},
\]
where $\lfloor\cdot\rfloor$ denotes the floor function.
For a given sample size $m_\sample$, the batch size is set to
\[
m_\batch=\max\left\{\left\lfloor 10\sqrt{m_\sample}\right\rfloor,\,1000\right\},
\]
and the maximum number of batches to
\[
T_\batch=\max\left\{20,\,\left\lfloor0.15\sqrt{m_\sample}\right\rfloor\right\}.
\]
\end{remark}

\section{Greedy starting point algorithm} \label{sec_gspa}
A crucial step in incremental clustering is the selection of a starting
center for the new cluster. Earlier nonsmooth incremental clustering methods
\cite{BagAliSUl2023,KarBagTah:2018,bigClust:2025,BagTahUgo:2016,
KarBagTah:2017,OrdBag:2015,BagOrdOztXav:2015} employ a multi-start strategy
in which several candidate starting points are generated, an auxiliary
clustering problem is solved from each candidate, and the resulting
centers are subsequently used to initialize the corresponding clustering problem, which is then solved multiple times. Although this strategy has been shown to produce high-quality
solutions, repeated auxiliary and clustering optimization runs may
become computationally expensive for very large datasets.

In \isicns, we replace this procedure with \gspans, which directly constructs
a single promising starting center $\bar{\y}\in\R^n$ for the new cluster.
The resulting center is appended
to the current cluster centers and used directly to initialize the next
clustering problem, thereby eliminating the need to solve 
either a separate auxiliary clustering problem or
the same clustering problem multiple times from different starting points.
%a separate auxiliary clustering problem. 

Assume that the solution $(\x_1,\ldots,\x_{l-1})$ to the $(l-1)$-clustering
problem is known for the dataset $A$. Here and throughout this section, $A$
denotes the dataset supplied to \gspans; when \gspa is used within \isic for
big datasets (e.g., $m>100\,000$), this may be the sample $A_\sample$ rather
than the complete dataset.  For each data point
$\ba_i\in A$, $i=1,\ldots,m$, let
$c(i)\in\{1,\ldots,l-1\}$ denote the index of its closest cluster center,
that is,
\[
c(i)\in
\argmin_{j=1,\ldots,l-1} \ d(\x_j,\ba_i).
\]
If several cluster centers are at the same minimum distance from $\ba_i$,
ties are broken arbitrarily so that each data point is assigned to exactly
one cluster.
The squared distance from $\ba_i$ to
its assigned cluster center is then
\begin{equation}
\label{rdistance_gspa}
r_{l-1}^i
=
d(\x_{c(i)},\ba_i)
=
\min_{j=1,\ldots,l-1}d(\x_j,\ba_i),
\end{equation}
and the corresponding clusters are
\[
A^j
=
\{\ba_i\in A:c(i)=j\},
\qquad
j=1,\ldots,l-1.
\]

If a candidate point $\bar{\ba}$ were added as a new cluster center while
the existing centers remained fixed, the contribution of a data point
$\ba_i$ to the clustering objective would change from $r_{l-1}^i$ to
\[
\min\big\{r_{l-1}^i,d(\bar{\ba},\ba_i)\big\}.
\]
Hence, the corresponding change in its objective contribution is
\[
\min\big\{0,d(\bar{\ba},\ba_i)-r_{l-1}^i\big\},
\]
which is nonpositive and is strictly negative whenever $\bar{\ba}$ is closer
to $\ba_i$ than its currently assigned cluster center.

Let $\bar{A}\subseteq A$ denote the candidate set formed by selecting the most
distant points within each non-trivial cluster (see lines \ref{line:gspa_candi_start}--\ref{line:gspa_candi_end} of Algorithm \ref{alg_find_sp_greedy}). For each candidate
$\bar{\ba}\in \bar{A}$, let
\[
\bar r_{l-1}
=
\min_{j=1,\ldots,l-1}d(\x_j,\bar{\ba})
\]
denote its squared distance to the closest existing cluster center.
Based on the potential decreases above, we define the
{\em improvement potential} of $\bar{\ba}$ as
{ %\color{red}
\[
\Phi(\bar{\ba})
=
\sum_{\ba_i\in A:\; r_{l-1}^i>\frac14\bar r_{l-1}}
\min\bigl\{0,\,
d(\bar{\ba},\ba_i)-r_{l-1}^i
\bigr\},
\]
where the condition $r_{l-1}^i>\frac14\bar r_{l-1}$ excludes data points that cannot contribute to the improvement potential, thereby avoiding unnecessary distance evaluations (see Remark~\ref{rem3}).}
Thus, $\Phi(\bar{\ba})$ measures the total potential decrease contributed by
the data points considered for the candidate $\bar{\ba}$. Since every term
in $\Phi$ is nonpositive, minimizing $\Phi$ selects the candidate with the
largest improvement potential. Accordingly,
\[
\ba^*
\in
\argmin_{\bar{\ba}\in \bar{A}}\ \Phi(\bar{\ba}).
\]
The final starting center is obtained by averaging the data points in a set $B$ consisting of observations that are sufficiently close to the selected candidate $\ba^*$ and would benefit from reassignment to it. This provides a more representative starting center $\bar{\y}$ than using $\ba^*$ alone.
{ %\color{red} 
The greedy starting-point procedure is presented in a simplified form in Algorithm~\ref{alg_find_sp_greedy}. A more detailed version, which allows better control and analysis of the computational complexity, is provided in Supplementary Material~1.}

\begin{algorithm*} % 
\caption{\gspans($A,(\x_1,\ldots,\x_{l-1}),l$)} %\hfill Simple version}
\label{alg_find_sp_greedy}
{\small
\begin{algorithmic}[1]
\Require Dataset $A=\{\ba_1,\ldots,\ba_m\}\subset\R^{n}$;
number of clusters $l\ge 2$;
solution $(\x_1,\ldots,\x_{l-1}) \in \R^{n(l-1)}$ to the $(l{-}1)$-clustering problem.

\Parameters Candidate-set size parameter $\bar{m}\in\{1,\ldots,m\}$;  improvement threshold $\eta\in(0,1]$.

\Ensure  A single starting center $\bar{\y}\in\R^n$ for the $l$-clustering problem.
 \For{$i=1,\ldots,m$}
 \Comment{Assignment and distance computation}
    \State $c(i)\leftarrow
           \argmin_{j=1,\ldots,l-1} d(\x_j,\ba_i)$
    \State $r^i_{l-1}\leftarrow d(\x_{c(i)},\ba_i)$
\EndFor
\State $\mathcal I_{\bar A}\leftarrow\emptyset$
    \Comment{Candidate indices; $\bar A=\{\ba_i:i\in\mathcal I_{\bar A}\}$}\label{line:gspa_candi_start}
\For{$j=1,\ldots,l-1$}
    \State $A^j\leftarrow\{\ba_i\in A : c(i)=j\}$,\quad
           $n_j\leftarrow|A^j|$ 
    \If{$n_j > 1$}
 \State $q_j\leftarrow
        \max\{1,\lfloor n_j\bar m/m\rfloor\}$
\State Select $I_j\subseteq\{i:c(i)=j\}$,
       $|I_j|=q_j$, corresponding to the
       largest $r^i_{l-1}$ in $A^j$
       %\State Select the indices $I_j$ of the $q_j$
       % points in $A^j$ with largest $r^i_{l-1}$
        \State $\mathcal I_{\bar A}\leftarrow
        \mathcal I_{\bar A}\cup I_j$
    \EndIf
\EndFor
\If{$\mathcal I_{\bar A}=\emptyset$}
    \State $\mathcal I_{\bar A}\leftarrow
    \bigl\{\argmax_{i=1,\ldots,m} r^i_{l-1}\bigr\}$\label{line:gspa_candi_end}
\EndIf

\For{each $\bar \imath\in\mathcal I_{\bar A}$}
    \Comment{Greedy choice of the candidate}
    \State $\bar{\ba}\leftarrow\ba_{\bar \imath}$,
           \quad $\bar r_{l-1}\leftarrow r_{l-1}^{\bar \imath}$
    \State $\Phi(\bar{\ba})\leftarrow 0$
           \Comment{Improvement potential}
    \For{$i=1,\ldots,m$}
        \If{$r^i_{l-1}>\frac14\bar r_{l-1}$}\label{line:gspa_screening}
            \State $\Phi(\bar{\ba})\leftarrow
            \Phi(\bar{\ba})+
            \min\{0,d(\bar{\ba},\ba_i)-r^i_{l-1}\}$
        \EndIf
    \EndFor
\EndFor

\State $\bar \imath^*\leftarrow
\argmin_{\bar \imath\in\mathcal I_{\bar A}}
\Phi(\ba_{\bar \imath})$
\State $\ba^*\leftarrow\ba_{\bar \imath^*}$,
       \quad $r_{l-1}^*\leftarrow r_{l-1}^{\bar \imath^*}$ \Comment{Best candidate selection}

\State $B\leftarrow\bigl\{\ba_i\in A : r^i_{l-1}> \frac{1}{4} \,{r}^*_{l-1}
       \;\text{and}\;
       d(\ba^*,\ba_i)<\eta\,r^i_{l-1}\bigr\}$ \Comment{Selection of attracted observations}\label{line:gspa_B}

\If{$B\neq\emptyset$} \Comment{Starting center}
\State $\bar{\y} \leftarrow \frac{1}{|B|}\sum_{\ba_i \in B} \ba_i$
\Else
\State $\bar{\y} \leftarrow \ba^*$
\EndIf
\State \Return $\bar{\y}$
\end{algorithmic}}
\end{algorithm*}

\begin{remark}[Parameter choices]\label{rem3}
{%\color{red}
Let $\bar{\ba}=\ba_{\bar \imath}$ be a candidate center, and let $\bar r_{l-1}=r_{l-1}^{\bar \imath}$ and $r_{l-1}^i$ denote the squared distances of $\bar{\ba}$ and $\ba_i$, respectively, from their nearest cluster centers. By the reverse triangle inequality,
$$
\begin{aligned}
\|\bar{\ba}-\ba_i\| \,
\geq \,\|\bar{\ba}-\x_{c(i)}\|
-\|\ba_i-\x_{c(i)}\|
\,\geq \,\sqrt{\bar r_{l-1}}-\sqrt{r_{l-1}^i}.
\end{aligned}
$$
Consequently, if $r_{l-1}^i\leq\frac{1}{4}\bar r_{l-1}$, then
$$
d(\bar{\ba},\ba_i)=\|\bar{\ba}-\ba_i\|^2
\geq r_{l-1}^i,
$$
and $\ba_i$ cannot contribute to the improvement potential of $\bar{\ba}$. 
This justifies the screening condition $r_{l-1}^i>\frac{1}{4}\bar r_{l-1}$ used when computing the improvement potential in line \ref{line:gspa_screening} of Algorithm~\ref{alg_find_sp_greedy}. The same argument applies to the selected candidate $\ba^*$, with $\bar r_{l-1}$ replaced by $r_{l-1}^*$, justifying the screening condition in the construction of $B$ (see line \ref{line:gspa_B} of Algorithm ~\ref{alg_find_sp_greedy}).
These results hold regardless of whether the data points share the same nearest cluster center.
}

The improvement threshold $\eta\in(0,1]$ controls how restrictive the
selection of points around $\ba^*$ is (see line \ref{line:gspa_B} of Algorithm ~\ref{alg_find_sp_greedy}). For $\eta=1$, all points that would
benefit from reassignment to $\ba^*$ may be included, even when the
improvement is small. Using $\eta<1$ introduces a margin, so that only points
for which $\ba^*$ provides a sufficiently pronounced improvement contribute
to the construction of the new starting center. This reduces the influence
of borderline points that may not be representative of the region associated
with the new cluster. 
In our implementation, the improvement threshold is fixed at $\eta=0.6$ for all experiments.

The parameter $\bar{m}$ in Algorithm~\ref{alg_find_sp_greedy} limits the number of candidate starting points considered when adding a new cluster, with the candidates distributed among the existing clusters in proportion to their sizes. As the number of clusters increases, evaluating candidate starting points becomes progressively more expensive. Thus, the choice of $\bar{m}$ reflects a computational trade-off rather than a theoretically prescribed value. In our implementation, the initial value is set to $\bar{m}=500$ when $mn\leq 10^7$ and to $\bar{m}=200$ otherwise. The value of $\bar{m}$ is then gradually decreased as the number of clusters increases, with corresponding lower bounds of $200$ and $100$, respectively.
\end{remark}

\section{Stochastic inexact LMBM} \label{sec_islmb}
The optimization engine of \isic is \islmbm --- a novel stochastic variant of \ilmbm \cite{inexactLMBM2026}. In its turn, \ilmbm is a limited-memory bundle method that is designed for large-scale NSO with inexact function values and subgradient information. This robustness to inexact information is particularly relevant in stochastic optimization, where quantities computed from batches provide only approximations of their full-data counterparts. By explicitly accounting for such inexactness, \islmbm enables \isic to better handle the inaccuracies introduced by stochastic computations. In addition, as in the underlying limited-memory bundle methodology \cite{HaaMieMak:2004,HaaMieMak:2007,inexactLMBM2026}, both the storage requirements and the computational cost of forming the search direction grow only linearly with the number of optimization variables.

For the finite-sum problem \eqref{sum-function}, \islmbm operates on the batch functions $f_\batch$ with $m_\batch \leq M$ instead of the whole objective $\Jobj$. At each outer iteration $t$, a batch $\batch_t$ of size $m_\batch$ is selected uniformly, and \ilmbm is applied to the corresponding batch problem for at most $i_{\max}$ iterations before the next batch is selected (inner iterations). Information generated during the optimization is retained through the limited-memory correction matrices $S^t$ and $U^t$ (see, e.g., \cite{HaaMieMak:2004,HaaMieMak:2007,ByrNocSch:1994}). These matrices store a small number, $\hat{m}_c$, of recent correction vectors formed from differences between successive iterates and the corresponding subgradient information. By carrying this information from one batch to the next, \islmbm can use information from previous batches when constructing subsequent search directions.

Algorithm~\ref{alg_SILMBM} gives the \islmbm procedure. For a detailed description and analysis of the underlying \ilmbm method, we refer to \cite{inexactLMBM2026}, while its algorithmic framework is summarized in Supplementary Material 1.

\begin{algorithm*} % 
\caption{\islmbmns($\x,f,m_\batch,T_\batch$)}
\label{alg_SILMBM}
{\small
\begin{algorithmic}[1]
\Require A starting point $\x\in\R^\dimopti$;
a finite-sum function $\Jobj(\x)=\frac{1}{M}\sum_{i=1}^{M}f_i(\x)$ to be minimized;
batch size $m_\batch\leq\numpart$;
maximum number of batches $T_\batch\geq1$.

\Parameters Iteration limit $i_{\max}>0$;
number of stored correction vectors $\hat{m}_c\geq3$.

\Ensure A heuristic minimizer $\x^*$ of $\min_{\x\in\R^\dimopti}f(\x)$.

\State $\x^1\leftarrow \x$, \quad $t=1$
    \Comment{Initialization}
\State $S^1 \gets (\bot, \dots, \bot) \in \left( \mathbb{R}^N \cup \{\bot\} \right)^{\hat{m}_c}$ \Comment{Initialize empty $\hat{m}_c$-length circular buffer for iterate differences}
\State $U^1 \gets (\bot, \dots, \bot) \in \left( \mathbb{R}^N \cup \{\bot\} \right)^{\hat{m}_c}$ \Comment{Initialize empty $\hat{m}_c$-length circular buffer for subgradient differences}

\While{$t \leq T_\batch$}
\State $\batch_{t} \leftarrow \Call{UniformSample}{\{1,\ldots,M\},m_{\batch}}$ 
    \Comment{Batch selection}
    
    \State $\x^{t+1},S^{t+1},U^{t+1} \leftarrow \Call{InexactLMBM}{\x^t,f_{\batch_t},S^t,U^t,\hat m_c,i_{\max}}, \; \text{where } f_{\batch_t}(\cdot)=\frac{1}{m_\batch}\sum_{i \in \batch_t} f_i(\cdot)$ \\
    \Comment{Solving the batch problem}
    
    \State $t \leftarrow t+1$
\EndWhile
\State \Return $\x^* \leftarrow \x^{t}$
\end{algorithmic}}
\end{algorithm*}

\begin{remark}\label{rem1}
Similarly to \slmbm used in \bigClust \cite{bigClust:2025}, \islmbm is heuristic and is not guaranteed, in general, to converge to a stationary point (see Supplementary Material 1) of the objective function $\Jobj$. However, when Algorithm \ref{alg_SILMBM} is used with $m_\batch=\numpart$ and a sufficiently large $i_{\max}$ (or $T_\batch$), it reduces to the original \ilmbmns, apart from a few additional iterations before termination. Consequently, for $m_\batch=\numpart$, Algorithm \ref{alg_SILMBM} inherits the global convergence properties of \ilmbm for LLC functions \cite{inexactLMBM2026} (see also Theorem S1 in Supplementary Material 1).
%Instead, Algorithm \ref{alg_SILMBM} terminates after $T_{\batch}$ batches; in our experiments, {\color{red}$T_{\batch}=10$}.\todo{tarkista arvo sekä klusteroinnille että aux klusteroinnille!}
\end{remark}

\begin{remark}
Unlike \slmbm in \cite{bigClust:2025}, which uses the best objective-function value computed over the full dataset as part of its stopping criterion, \islmbm terminates after a prescribed maximum number of batches $T_\batch \geq 1$. This choice avoids repeated full-data evaluations of the clustering objective, which can be computationally expensive for very large datasets. It also makes the algorithm more readily applicable to settings where the complete dataset is not simultaneously available or may change over time, such as streaming or dynamic data. 
\end{remark}

\begin{remark}[Time and memory complexity]\label{ordo}
Let $m_\sample$ denote the sample size used by \isic and $n$ the number of features in the data. The working-memory requirement of \isic is $O(m_\sample n)$, assuming that the limited-memory parameter $\hat m_c$ is fixed and the maximun number of clusters $k<m_\sample$. Similarly, since $m_\batch\leq m_\sample$, for fixed $\bar{m}$, $T_\batch$, $i_{\max}$, and $k$, the computational cost of \isic is
$O(m_\sample n)$.  Thus, both the working-memory requirement and computational
cost scale linearly with the sample size and the number of features. 
{A more detailed complexity analysis is provided in Supplementary Material~1.}
%Thus, the working-memory requirement grows only linearly with both the sample size and the number of features. 

If the complete dataset is stored in memory, its storage additionally requires $O(mn)$ memory; however, as discussed above, \isic does not necessarily require the complete dataset to be simultaneously available in memory.

\end{remark}

\section{Numerical experiments}\label{sec_num}

In this section, we compare the performance of the proposed \isic algorithm with three other algorithms for large and big data clustering: namely, \sclust \cite{bigClust:2025}, \lmbclust \cite{KarBagTah:2018}, and \bigmeans \cite{MusMlaJarMus2023}. These three algorithms were selected based on their performance in \cite{bigClust:2025}, and the experimental setup is kept similar to that study to facilitate comparison and highlight the differences and new contributions of this work. Detailed results are reported in Supplementary Material 2, together with additional comparison with five other large and big data clustering algorithms, specifically, \hgmeans \cite{GriVid2019}, \drsmeans \cite{KozMus2024}, \dcclust \cite{BagTahUgo:2016}, \msmgkm \cite{OrdBag:2015}, and \mbkm from Scikit-learn \cite{scikit}.

\paragraph{Datasets}
The performance of \isic and other competitive algorithms is assessed using 22 publicly available real-world datasets. % considered also in \cite{bigClust:2025}. 
Table \ref{data} summarizes the main characteristics of these datasets, while more detailed information is available on the corresponding websites listed in Table \ref{data2}. All datasets consist only of numerical features and contain no missing values. The number of features ($n$) ranges from very few (2) to large (5\,000), and the number of data points ($m$) spans from thousands (the smallest 7\,797) to over ten million (the largest 10\,500\,000). For the numerical analysis, the datasets are divided into two equally sized groups according to their size: 11 datasets with fewer than 100\,000 samples and 11 datasets with more than 100\,000 samples.

\begin{table}[h!]
  \caption{\label{data} Brief description of the used datasets}
\small
  \begin{adjustbox}{max width=0.99\textwidth}
  %\resizebox{0.99\textwidth}{!}{
  \begin{tabular}{l c r c r c r } 
\hline\noalign{\smallskip}
  Dataset      &\phantom{}& $m$            &\phantom{}& $n$ &\phantom{}& $m \times n$  \\ %&\phantom{}& \\ % Ref. \\ %\midrule
\noalign{\smallskip}\hline\noalign{\smallskip}
  ISOLET                            &&       7\,797 &&  617 &&  4\,810\,749  \\ %&& \\ %\cite{Lic:2013}\\ % 11
  Gisette                           &&      13\,500 && 5\,000 && 67\,500\,000  \\ %&& \\ % \cite{Gisette}\\ % 14
  Gas Sensor Array Drift            &&      13\,910 &&  128 &&  1\,780\,480  \\ %&& \\ % \cite{GasSensor}\\ % 1
  EEG Eye State                     &&      14\,980 &&   14 &&   209\,720  \\ %&& \\ % \cite{Lic:2013} \\ % 2
  D15112                            &&      15\,112 &&    2 &&    30\,224  \\ %&& \\ % \cite{tsplib} \\ % 3
  Online News Popularity            &&      39\,644 &&   58 &&  2\,299\,352  \\ %&& \\ % \cite{OnlineNewsPopularity} \\ % 12
  KEGG Metabolic                    &&      53\,413 &&   20 &&  1\,068\,260  \\ %&& \\ % \cite{KEGGMetabolic}\\ % 4
  Shuttle Control               &&      58\,000 &&    9 &&   522\,000  \\ %&& \\ % \cite{Lic:2013}\\%$^*$\\ % 5
  Sensorless Drive Diagnosis        &&      58\,509 &&   48 &&  2\,808\,432  \\ %&& \\ % \cite{Lic:2013}\\ % 9
  MFCCs for Speech Emotion Recognition &&   85\,134 &&   58 &&  4\,937\,772  \\ %&& \\ % {\red \cite{} \\
  Pla85900                          &&      85\,900 &&    2 &&   171\,800  \\ %&& \\ % \cite{tsplib}\\ % 6
  %väliviiva
  \noalign{\smallskip} \hline \noalign{\smallskip}
  Music Analysis &&     106\,574 &&   518 &&  55\,205\,332 \\ %&& \\ % \cite{Lic:2013}\\ % 13
  MiniBooNE Particle Identification &&     130\,064 &&   50 &&  6\,503\,200  \\ %&& \\ % \cite{Lic:2013}\\ % 13
  Protein Homology                  &&     145\,751 &&   74 && 10\,785\,574  \\ %&& \\ % \\
  Range Queries Aggregates          &&     200\,000 &&    7 && 1\,400\,000 \\
  Skin Segmentation                 &&     245\,057 &&    3 &&   735\,171  \\ %&& \\ % \cite{SkinSegmentation}\\ % 7
  3D Road Network                   &&     434\,874 &&    3 &&  1\,304\,622  \\ %&& \\ % \cite{3DRoadNetwork} \\ % 8 
  Covertype                         &&     581\,012 &&   10 &&  5\,810\,120  \\ %&& \\ % \cite{Lic:2013}\\ % 15
  CORD-19 Embeddings$^*$            &&    1\,056\,660 &&  768 && 811\,514\,880  \\ %&& \\ % {\red \cite{}} \\ 
  US Census Data 1990               &&    2\,458\,285 &&   68 && 167\,163\,380  \\ %&& \\ % \cite{Lic:2013}\\ 
  BitcoinHeist                      &&	2\,916\,697	&&    8	&& 23\,333\,576  \\ %&& \\ % \\
  HEPMASS                           &&    10\,500\,000 &&  28 && 294\,000\,000  \\ %&& \\ % \cite{Lic:2013}\\ 
\noalign{\smallskip}\hline\noalign{\smallskip}
\multicolumn{7}{l}{\scriptsize $^*$CORD-19 Embeddings is continuously updated; we use the same version as in \cite{bigClust:2025}.}
\end{tabular}
\end{adjustbox}
\end{table}

\begin{table}[h!]
  \caption{\label{data2} Information about the used datasets (date accessed 24.09.2026)}
    \resizebox{1.0\textwidth}{!}{
 \begin{tabular}{l c l} 
\hline\noalign{\smallskip}
  Dataset      &\phantom{}& URLs             \\ 
\noalign{\smallskip}\hline\noalign{\smallskip}
  ISOLET                            && \url{https://archive.ics.uci.edu/dataset/54/isolet} \\ 
  Gisette                           && \url{https://archive.ics.uci.edu/dataset/170/gisette} \\ 
  Gas Sensor Array Drift            &&    \url{https://archive.ics.uci.edu/dataset/224/gas+sensor+array+drift+dataset} \\ 
  EEG Eye State                     &&    \url{https://archive.ics.uci.edu/dataset/264/eeg+eye+state} \\ 
  D15112                            &&    \url{https://github.com/mastqe/tsplib/blob/master/d15112.tsp}  \\ 
  Online News Popularity            &&    \url{https://archive.ics.uci.edu/dataset/332/online+news+popularity}  \\ 
  KEGG Metabolic                    &&    \url{https://archive.ics.uci.edu/dataset/220/kegg+metabolic+relation+network+directed}  \\ 
  Shuttle Control               &&    \url{https://archive.ics.uci.edu/dataset/148/statlog+shuttle}  \\ 
  Sensorless Drive Diagnosis        &&    \url{https://archive.ics.uci.edu/dataset/325/dataset+for+sensorless+drive+diagnosis}  \\ 
  MFCCs for Speech  &&  \\  
  \qquad Emotion Recognition && \url{https://www.kaggle.com/datasets/cracc97/features} \\ 
  Pla85900                          &&    \url{https://github.com/mastqe/tsplib/blob/master/pla85900.tsp}  \\ 
  %väliviiva
  \noalign{\smallskip} \hline \noalign{\smallskip}
   Music Analysis && \url{https://archive.ics.uci.edu/dataset/386/fma+a+dataset+for+music+analysis}\\
    MiniBooNE Particle &&     \\ 
  \qquad Identification &&    \url{https://archive.ics.uci.edu/dataset/199/miniboone+particle+identification}  \\ 
  Protein Homology                  &&    \url{https://www.kdd.org/kdd-cup/view/kdd-cup-2004/Data}  \\
  Range Queries Aggregates         && \url{https://archive.ics.uci.edu/dataset/493/query+analytics+workloads+dataset} \\
    Skin Segmentation                 &&    \url{https://archive.ics.uci.edu/dataset/229/skin+segmentation} \\ 
  3D Road Network                   &&    \url{https://archive.ics.uci.edu/dataset/246/3d+road+network+north+jutland+denmark} \\ 
  Covertype                         &&    \url{https://archive.ics.uci.edu/dataset/31/covertype} \\ 
  CORD-19 Embeddings                &&    \url{https://www.kaggle.com/datasets/allen-institute-for-ai/CORD-19-research-challenge}  \\ 
  US Census Data 1990               &&    \url{https://archive.ics.uci.edu/dataset/116/us+census+data+1990} \\ 
  BitcoinHeist                      && \url{https://archive.ics.uci.edu/dataset/526/bitcoinheistransomwareaddressdataset} \\
  HEPMASS                           &&    \url{https://archive.ics.uci.edu/dataset/347/hepmass} \\ 
\noalign{\smallskip}\hline\noalign{\smallskip}
\end{tabular}
}
\end{table}

\paragraph{Methods and setup}

The four methods, \isicns, \sclustns, \lmbclustns, and \bigmeansns, considered in the comparison are summarized in  Tables \ref{table:methods} and \ref{table:methods2}. The tables provide detailed information on their implementations, parameter settings, and other characteristics. Since general default parameter settings are not specified for \bigmeans in \cite{MusMlaJarMus2023}, we use parameter settings described in \cite{bigClust:2025}, including the version with inner parallelism using eight threads. For data sets with more than 100\,000 data points, we additionally consider versions of \isic with reduced sample and batch sizes, denoted by \isicns-\#, where $\# = 10, 20, 50$ (see Remark \ref{rem:isic_parameters}). Similarly, for \sclustns, we consider a reduced-data version denoted by \sclustns+. The parameter settings for \sclustns+ are those described in \cite{bigClust:2025}. In addition, for the optimization engine \islmbm of the proposed \isic algorithm, we set $i_{\max}=10$, while the default parameters of \ilmbm \cite{inexactLMBM2026} are used otherwise. The detailed parameter values used for all tested methods are provided in Supplementary Material 2.

\begin{table}[h!]
  \caption{\label{table:methods} Implementation details of the methods used for comparison}
  \resizebox{0.99\textwidth}{!}{
  \begin{tabular}{l l l c} 
\hline\noalign{\smallskip}
  Method & Repository & Parameters & Reference \\
\noalign{\smallskip}\hline\noalign{\smallskip}                        
\isic (Algorithm \ref{algorithm_isic}) & \url{https://github.com/napsu/ISIC} & Default defined in & -- \\
 &  & \quad Remarks \ref{rem:isic_parameters} and \ref{rem3} &  \\
\sclust & \url{https://github.com/napsu/bigClust} & Default & \cite{bigClust:2025} \\
\lmbclust & \url{https://github.com/napsu/LMB-Clust} & Default & \cite{KarBagTah:2018} \\
\bigmeans & \url{https://github.com/R-Mussabayev/bigmeans} & Defined in \cite{bigClust:2025} & \cite{MusMlaJarMus2023} \\
\noalign{\smallskip}\hline\noalign{\smallskip}
\end{tabular}
}
\end{table}

\begin{table}[h!]
  \caption{\label{table:methods2} Characteristics of the methods used for comparison}
  \resizebox{0.99\textwidth}{!}{
  \begin{tabular}{l l c l c l} 
\hline\noalign{\smallskip}
  Method & Approach & Incremental & Starting point strategy & Stochastic & Optimization engine \\
\noalign{\smallskip}\hline\noalign{\smallskip}  
\isic & NSO & Yes & \gspa (Algorithm \ref{alg_find_sp_greedy}) & Yes & \islmbm (Algorithm \ref{alg_SILMBM})\\
\sclust & NSO & Yes & Multistart \& auxiliary problem & Yes & \slmbm \\
\lmbclust & NSO & Yes & Multistart \& auxiliary problem & No & \LMBM \\
\bigmeans & $k$-means-based & No & $k$-means++ \& incumbent reuse across samples & Yes & $k$-means \\
\noalign{\smallskip}\hline\noalign{\smallskip}
\end{tabular}
}
\end{table}

The \isicns, \sclustns, and \lmbclust algorithms use the incremental approach to solve the clustering problems. Therefore, we compute up to 25 clusters with these methods, and the results with less than 25 clusters are readily available. Since \bigmeans does not give any intermediate results, we perform separate runs with different numbers of clusters, $k \in \mathcal{K} =\{2,3,4,5,10,15,20,25\}$, for comparison purposes. For the stochastic methods, we perform ten independent runs for each dataset and report the averaged results, while the non-stochastic algorithm \lmbclust requires only one run per dataset. Further, for \bigmeansns, the sample size is selected using a procedure similar to that in \cite{bigClust:2025}: two alternative sample-size settings are evaluated, and the one yielding the lower average cluster function value over ten runs is selected. Consequently, \bigmeans requires a total of 20 runs for each dataset.

The \isicns, \sclustns, and \lmbclust algorithms are implemented in Fortran 95, while \bigmeans is implemented in Python. To ensure comparable computational times across the different implementations, the running time of \bigmeans is measured using Python's {\tt time.time()} function, corresponding to the {\tt cpu\_time()} subroutine used in the Fortran implementations.

All computational experiments are carried out on iMac, 4.0 GHz Quad-Core Intel(R) Core(TM) i7 machine with 16 GB of RAM. We use gfortran to compile the Fortran codes and Python 3.10 with NumPy 1.24.2 and Numba 0.59.0 for \bigmeansns.

\paragraph{Evaluation metrics and results}

\begin{figure}[ht!] 
\centering
  \subfigure[Average error]{
   \includegraphics[width=\textwidth]{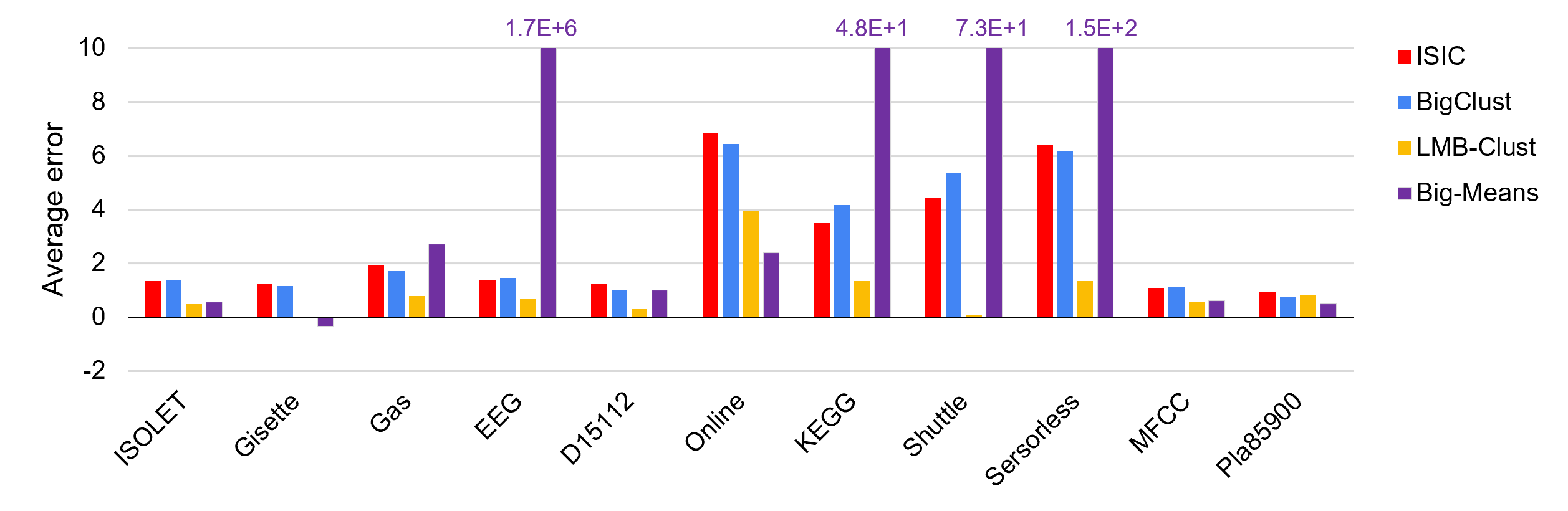}\label{fig:sde}} 
   \subfigure[Time for 10 clusters]{
   \includegraphics[width=\textwidth]{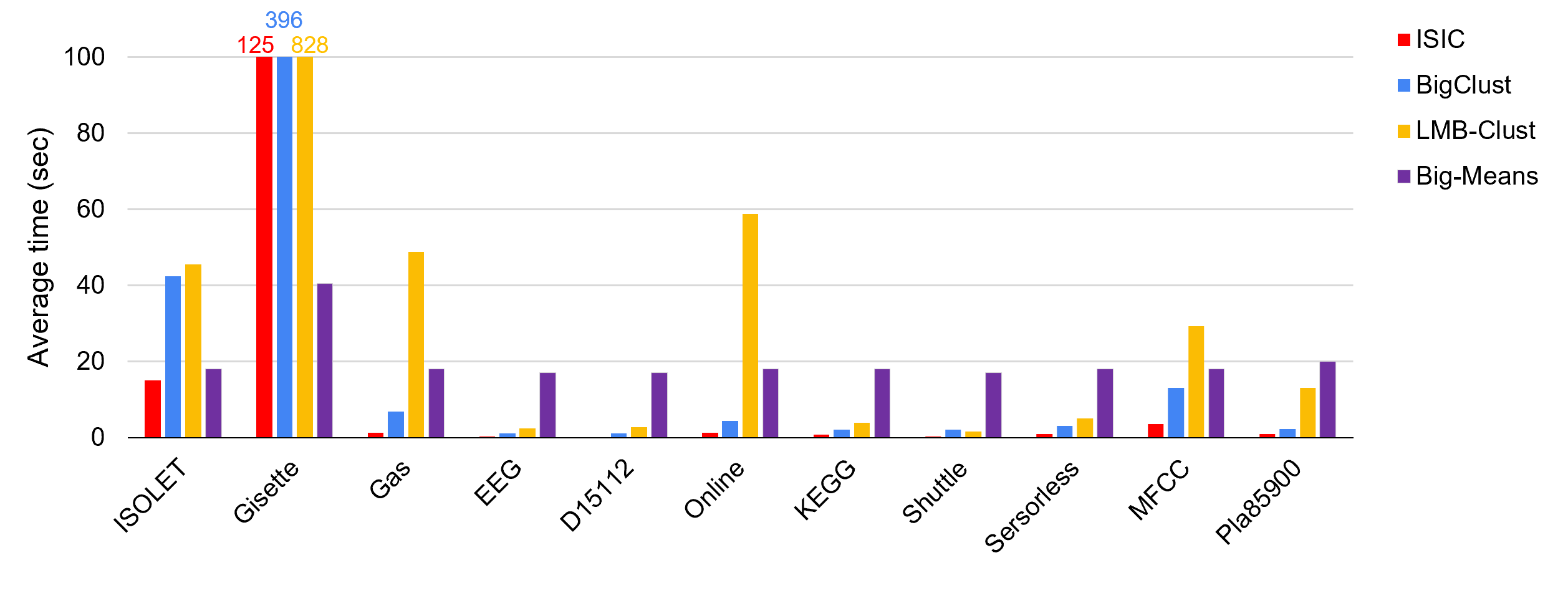}\label{fig:sdt10}}
  \subfigure[Time for 25 clusters]{
   \includegraphics[width=\textwidth]{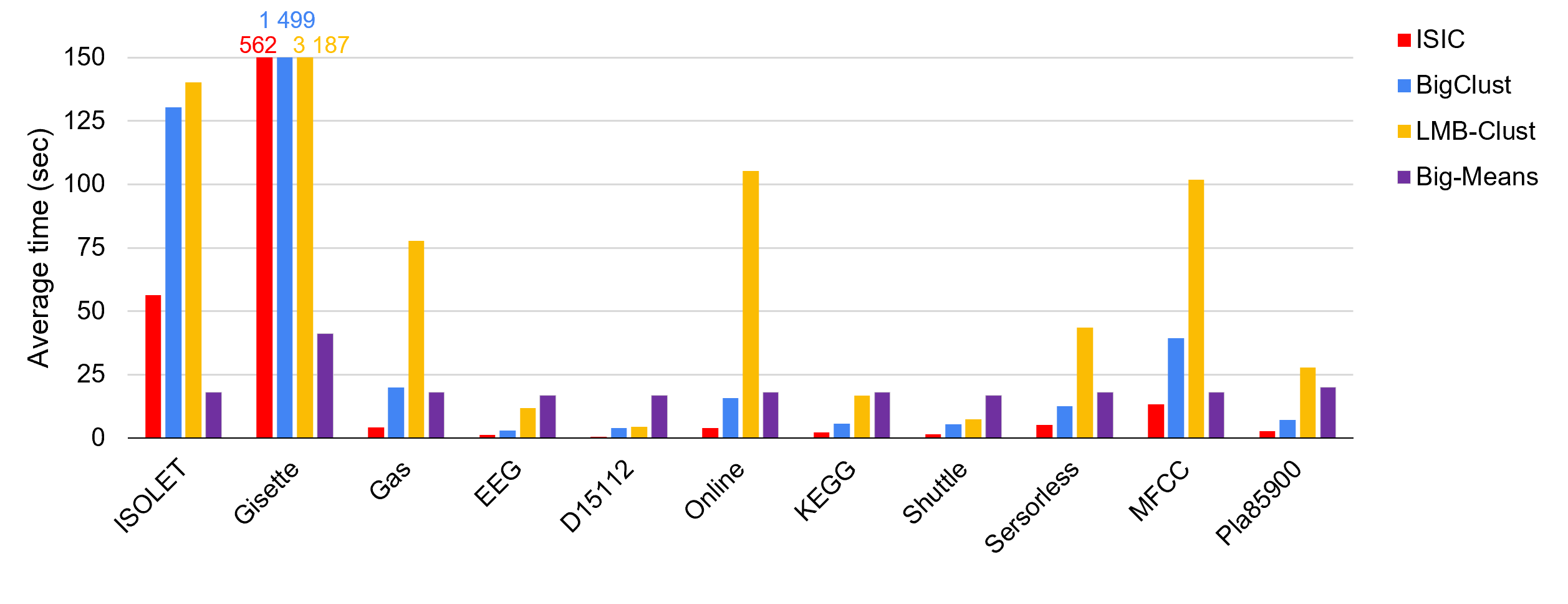}\label{fig:sdt25}}
\caption{Summary of results for datasets containing less than 100\,000 samples.}
\label{fig_summary1}
\end{figure}

\begin{figure}[ht!]
\centering
  \subfigure[Average error]{
   \includegraphics[width=\textwidth]{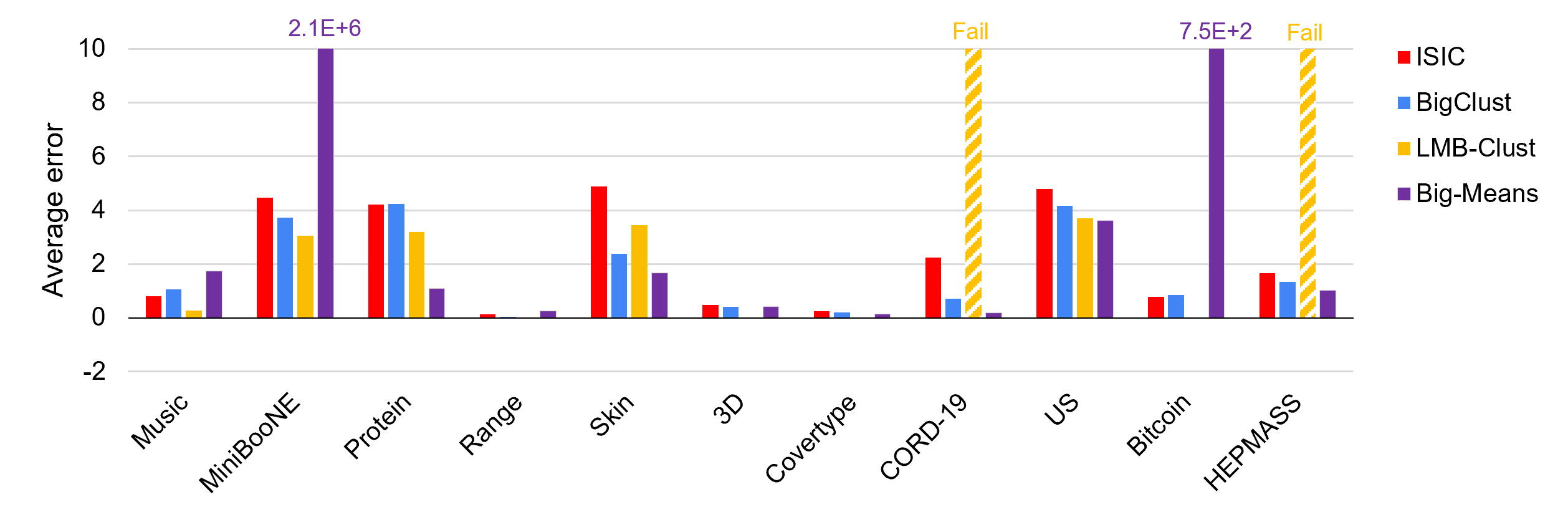}\label{fig:lde}} 
   \subfigure[Time for 10 clusters]{
   \includegraphics[width=\textwidth]{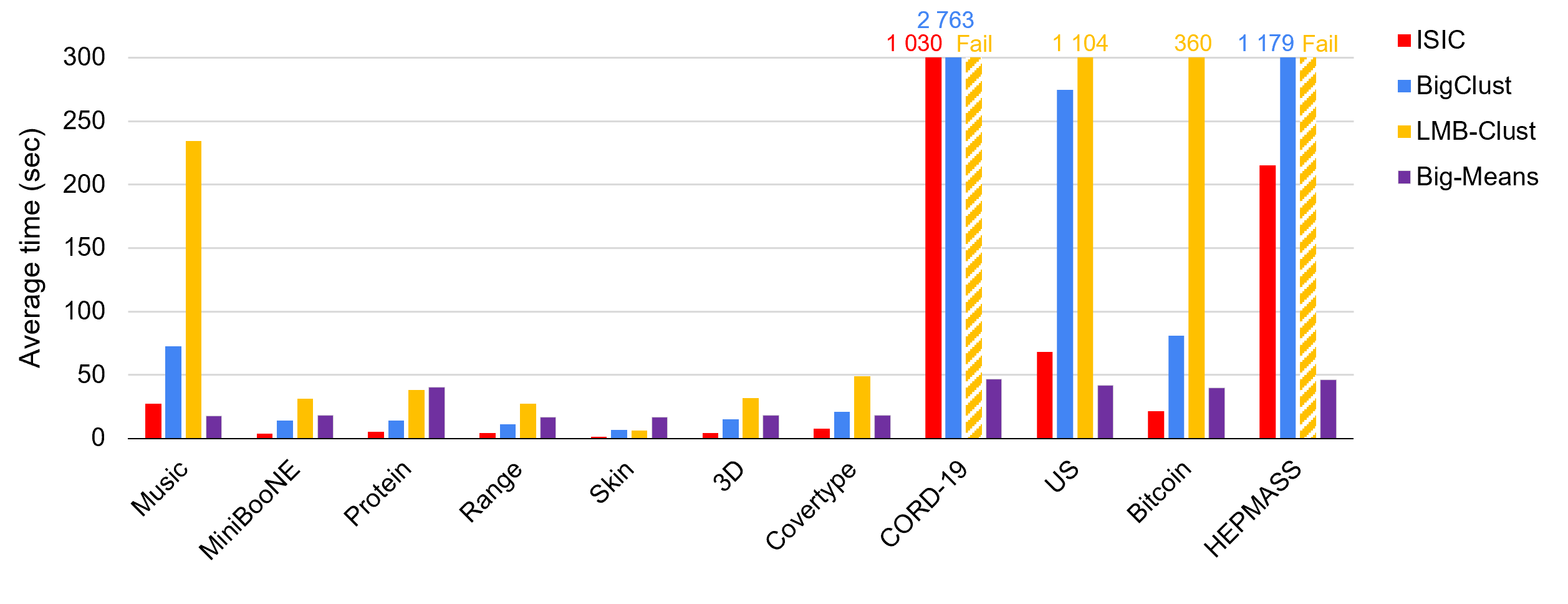}\label{fig:ldt10}}
  \subfigure[Time for 25 clusters]{
   \includegraphics[width=\textwidth]{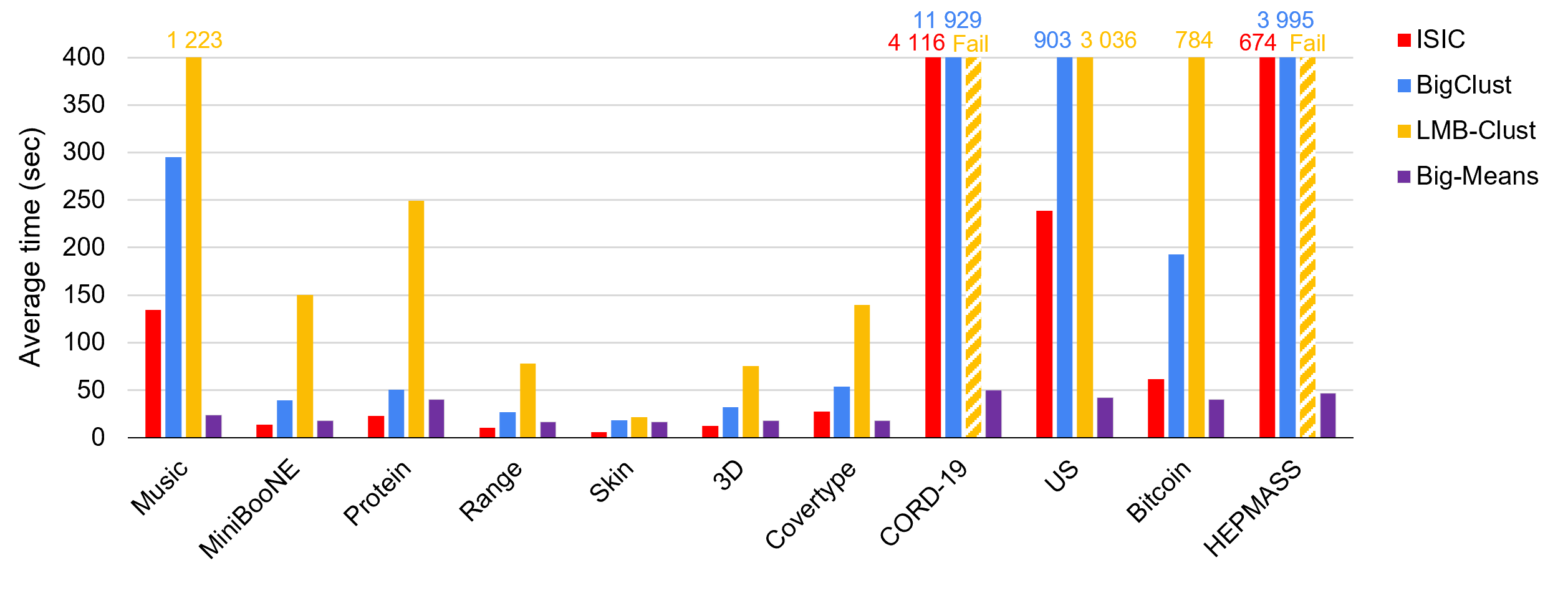}\label{fig:ldt25}}
\caption{Summary of results for datasets containing more than 100\,000 samples. Note: \lmbclust failed on the HEPMASS and CORD-19 Embeddings datasets because of memory limitations.}
\label{fig_summary2}
\end{figure}

\begin{figure}[h!]
  \centering
  \includegraphics[width=0.6\textwidth]{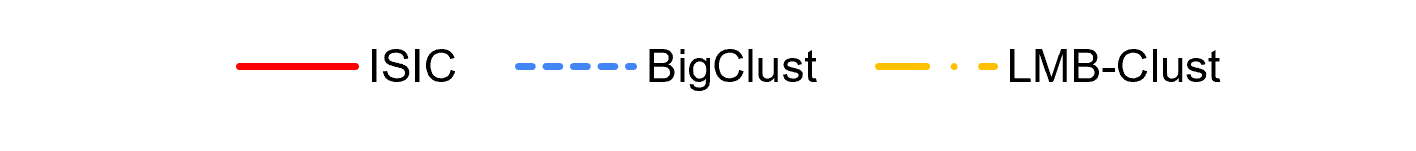}\\
\vspace{-3ex}
   \subfigure[Online News Popularity]{
   \includegraphics[width=0.48\textwidth]{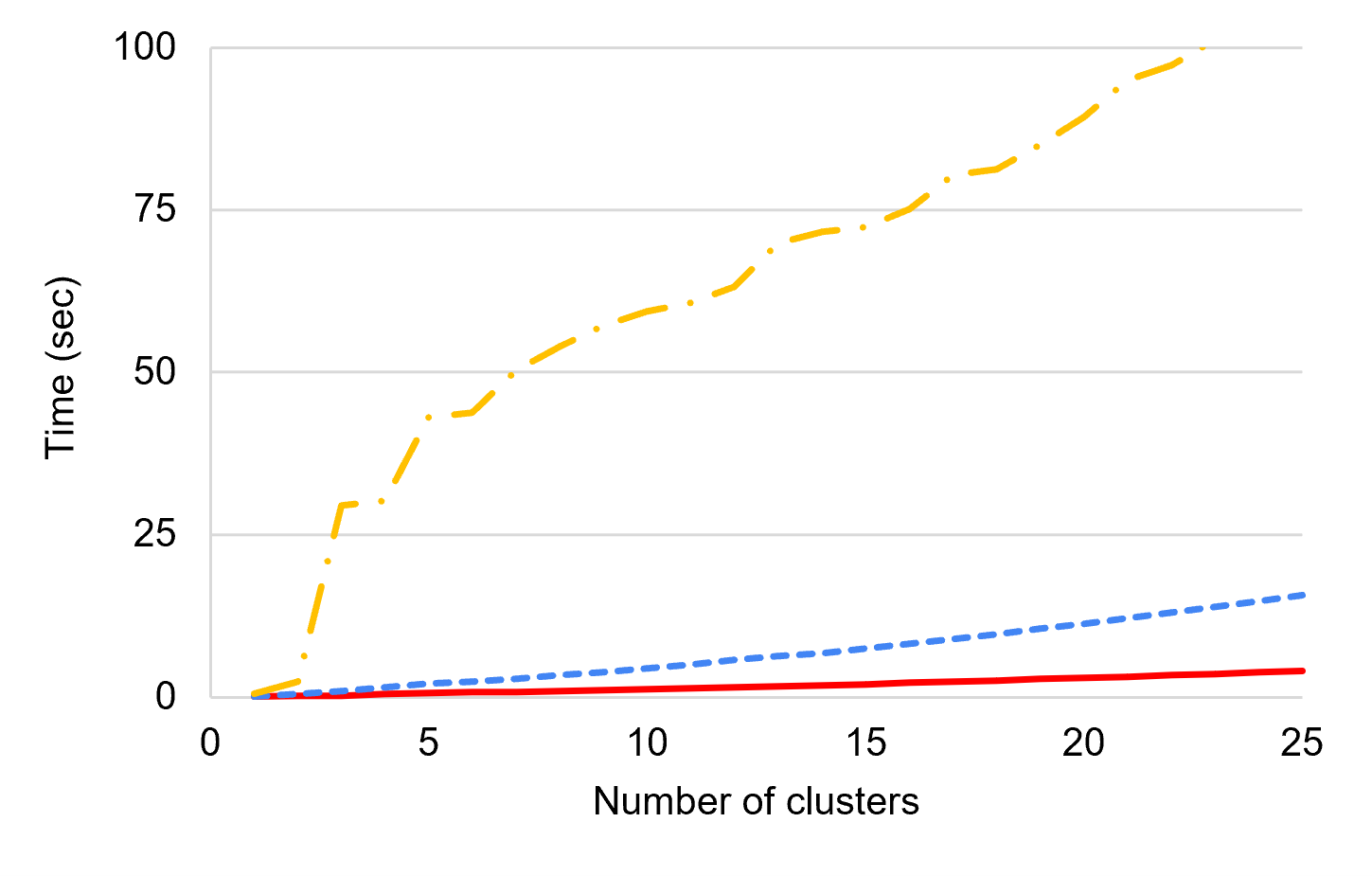}\label{fig:online-time}}
   \subfigure[Music Analysis]{
   \includegraphics[width=0.48\textwidth]{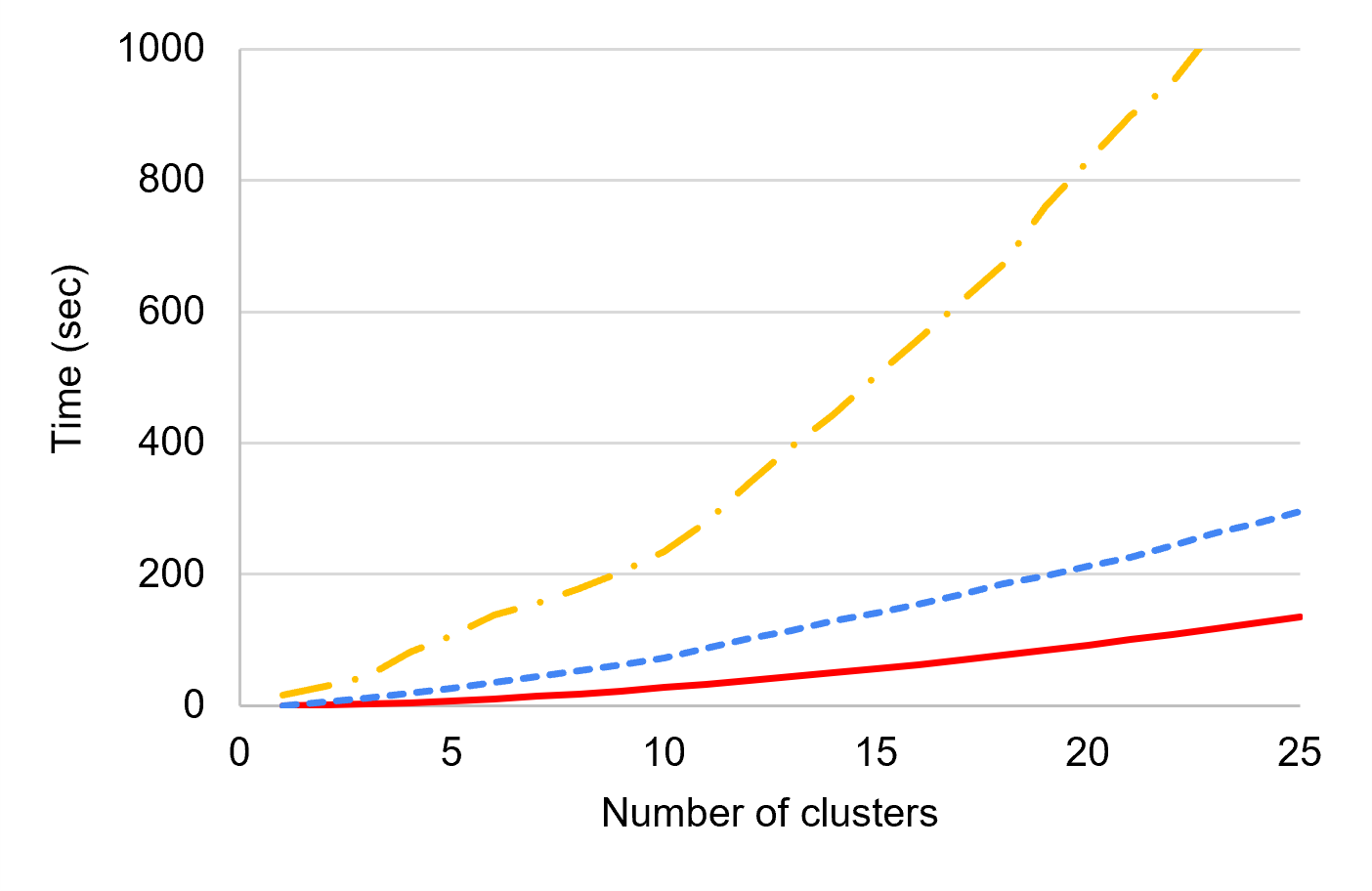}\label{fig:music-time}} 
\caption{Computational time versus number of clusters in datasets with different numbers of features: in Online News Popularity $m=39\,644$ and $n=58$, while in Music Analysis $m=106\,574$ and $n=518$.}
\label{fig_times}
\end{figure}

\begin{figure*}[h!]
  \centering
  \includegraphics[width=0.6\textwidth]{Kuvat/legend3.png}\\
\vspace{-3ex}
   \subfigure[Davies-Bouldin index]{
   \includegraphics[width=0.48\textwidth]{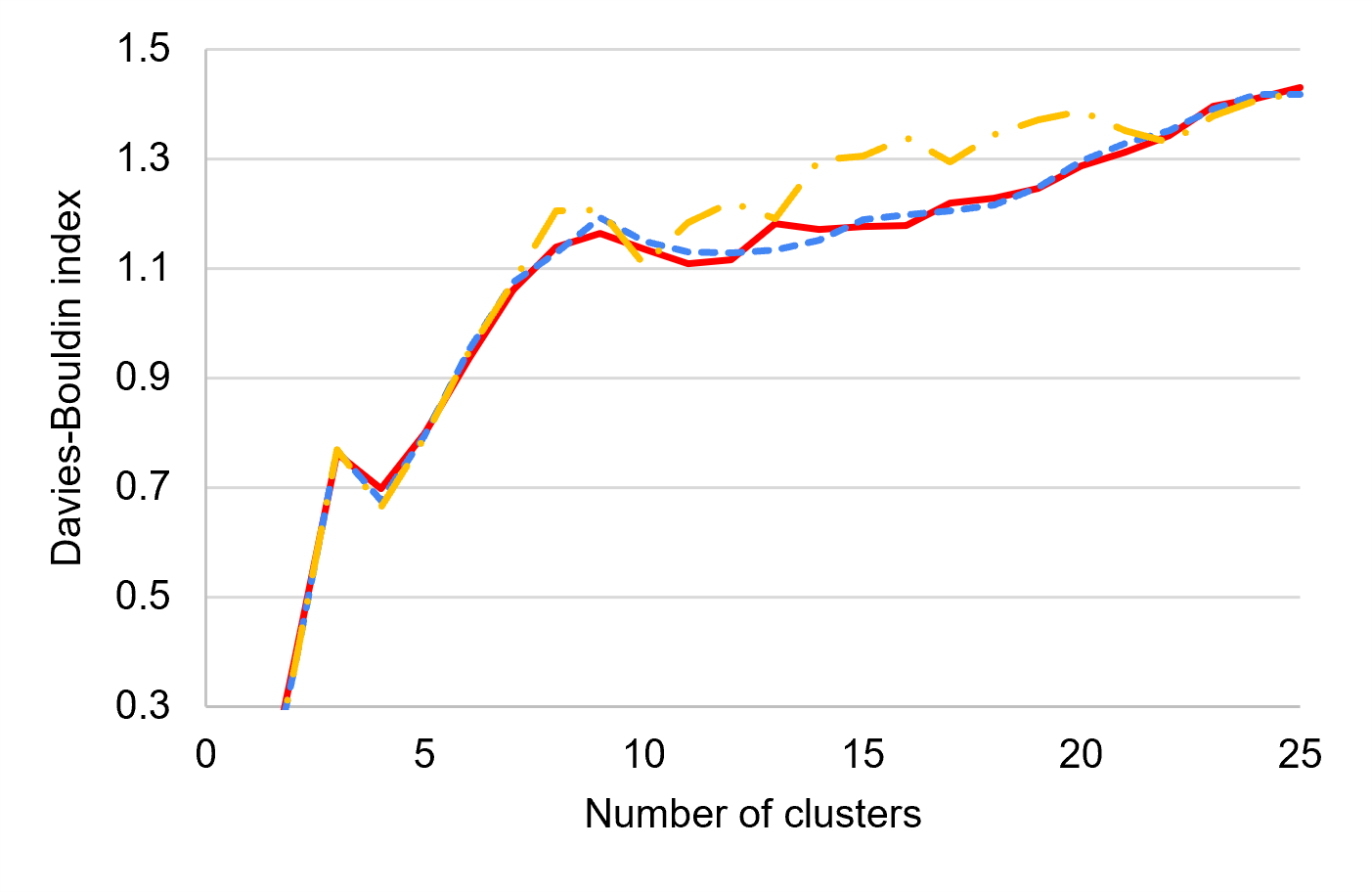}\label{fig:prothomo-Davis}}
  \subfigure[Dunn index]{
   \includegraphics[width=0.48\textwidth]{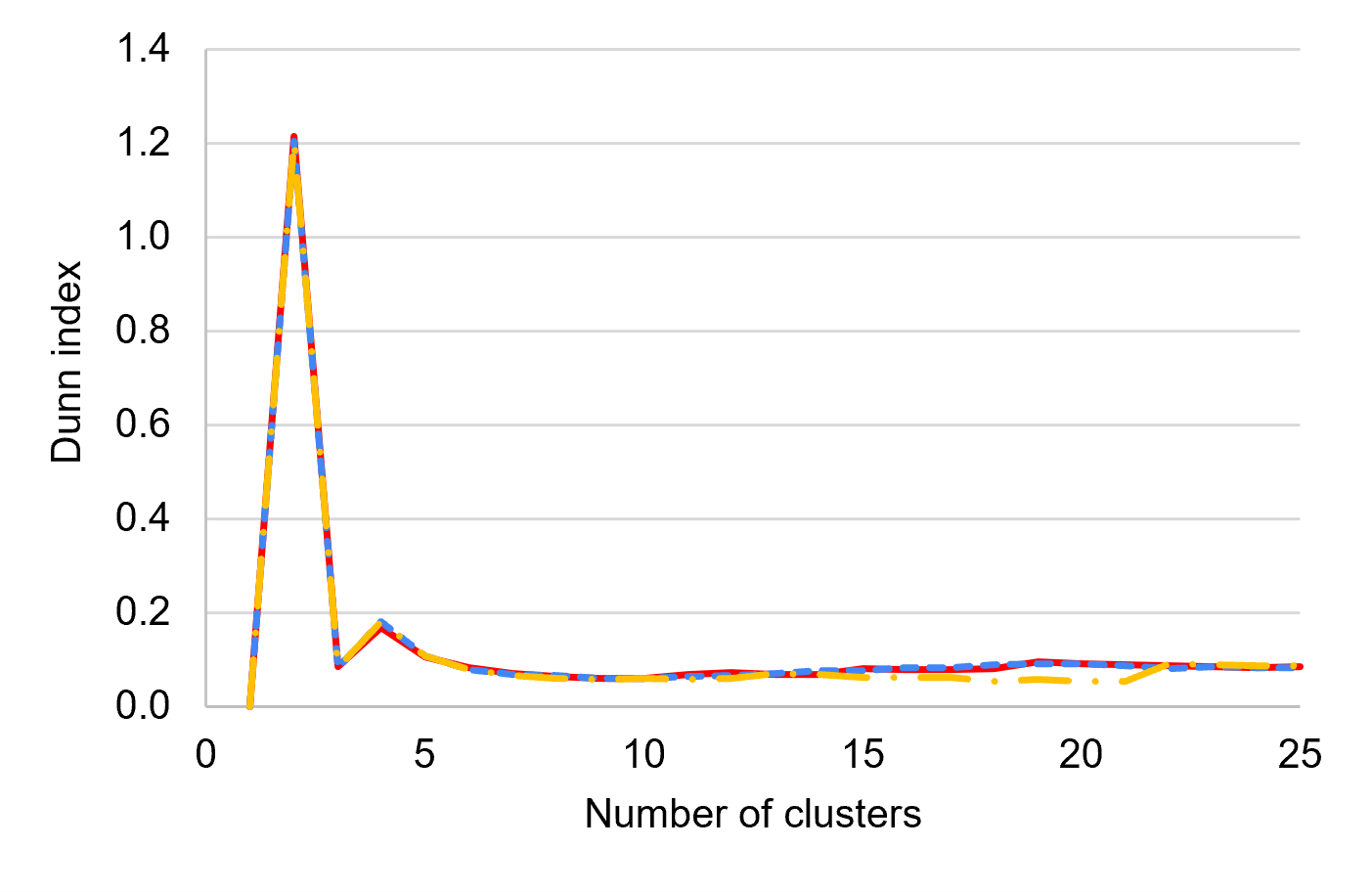}\label{fig:prothomo-Dunn}}
\caption{Protein Homology: Davies-Bouldin and Dunn validity indices vs.\ number of clusters.}
\label{fig_dbi_and_di}
\end{figure*}

\begin{figure}[h!]
\centering
\includegraphics[width=0.48\textwidth]{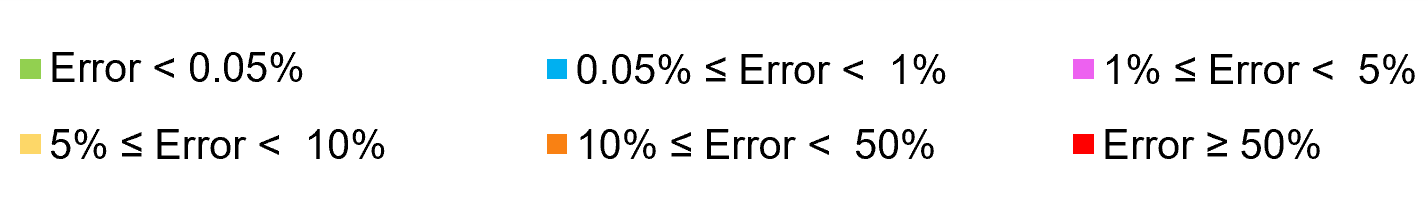}\\
\vspace{-2.5ex}
  \subfigure[]{
   \includegraphics[width=0.48\textwidth]{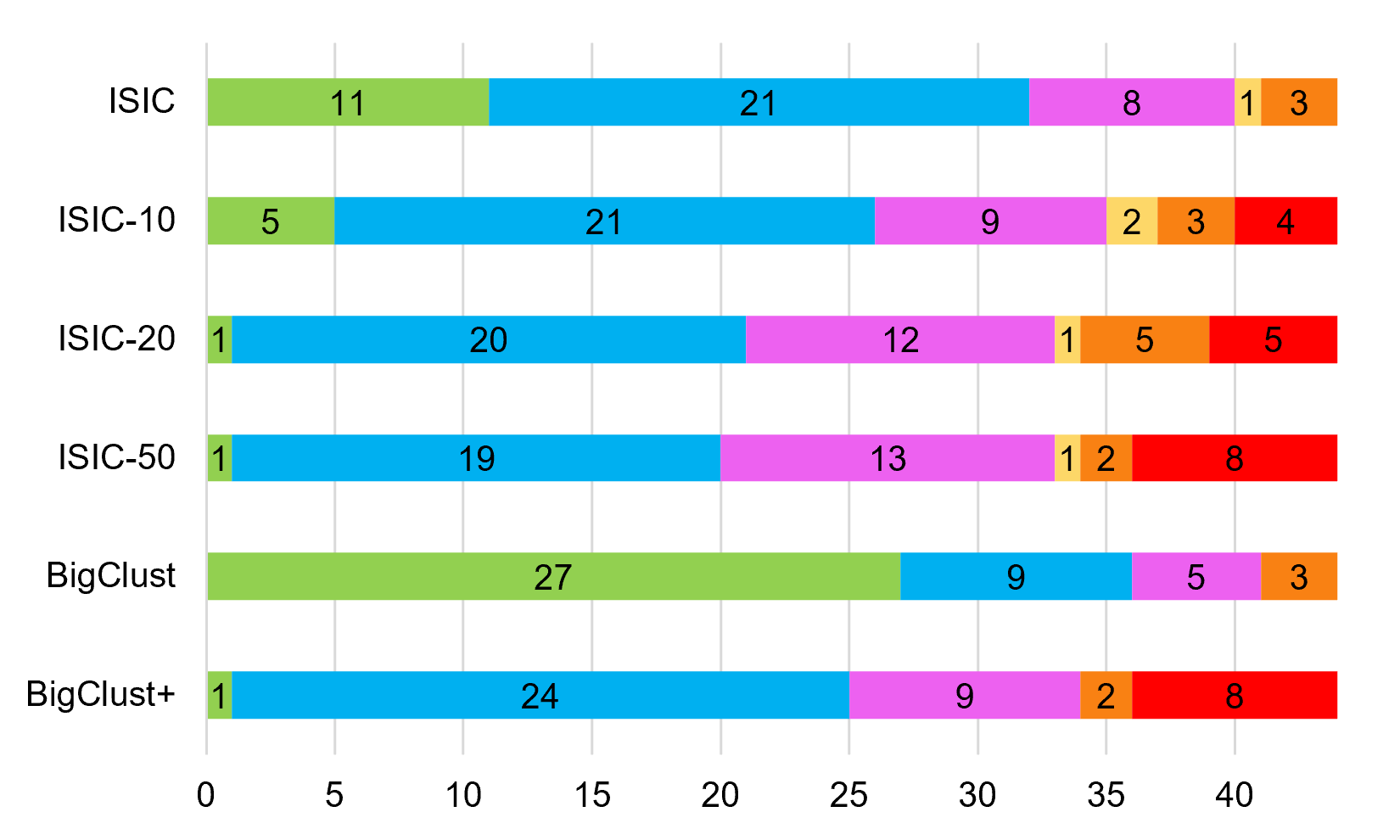}\label{fig:error_small}} 
   \subfigure[]{
   \includegraphics[width=0.48\textwidth]{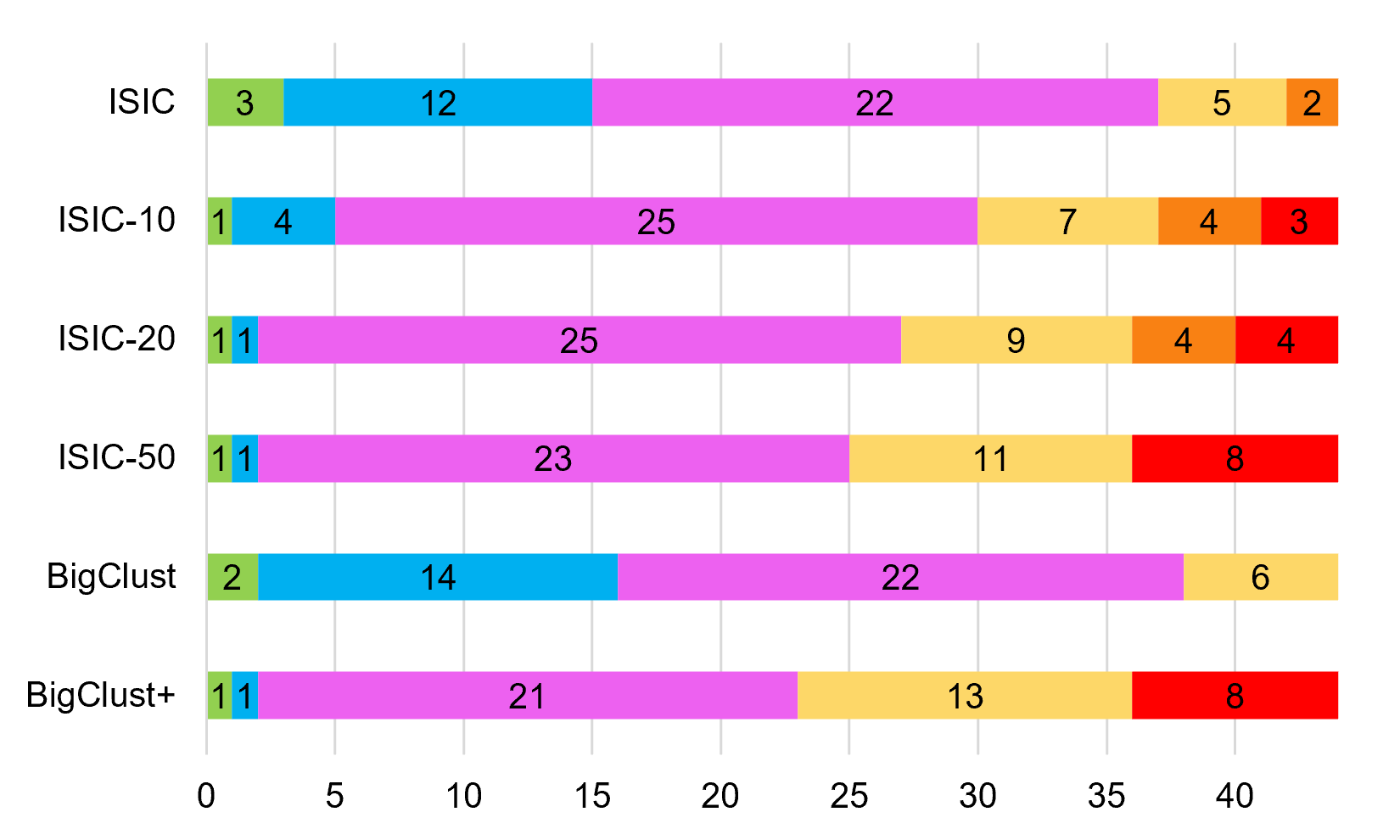}\label{fig:error_large}}
\caption{Cases categorized by relative errors for (a) $k$-clustering problems with $k=2,3,4,5$ and (b) $k$-clustering problems with $k=10,15,20,25$.}
\label{fig_error_subsample}\vspace{-0.25cm}
\end{figure}

\begin{figure}[h!]
\centering
\includegraphics[width=0.47\textwidth]{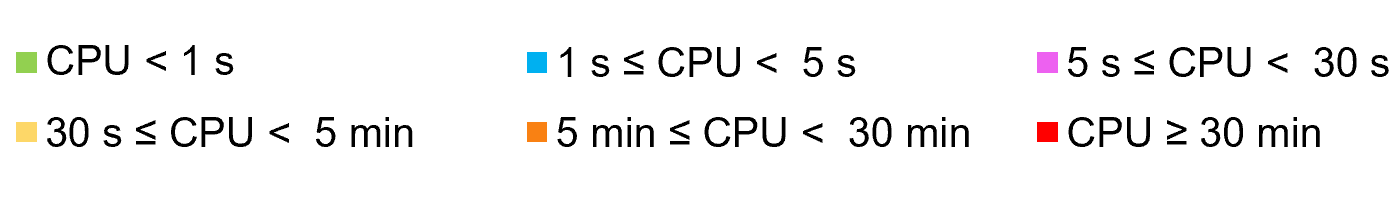}\\
\vspace{-2.5ex}
  \subfigure[]{
   \includegraphics[width=0.48\textwidth]{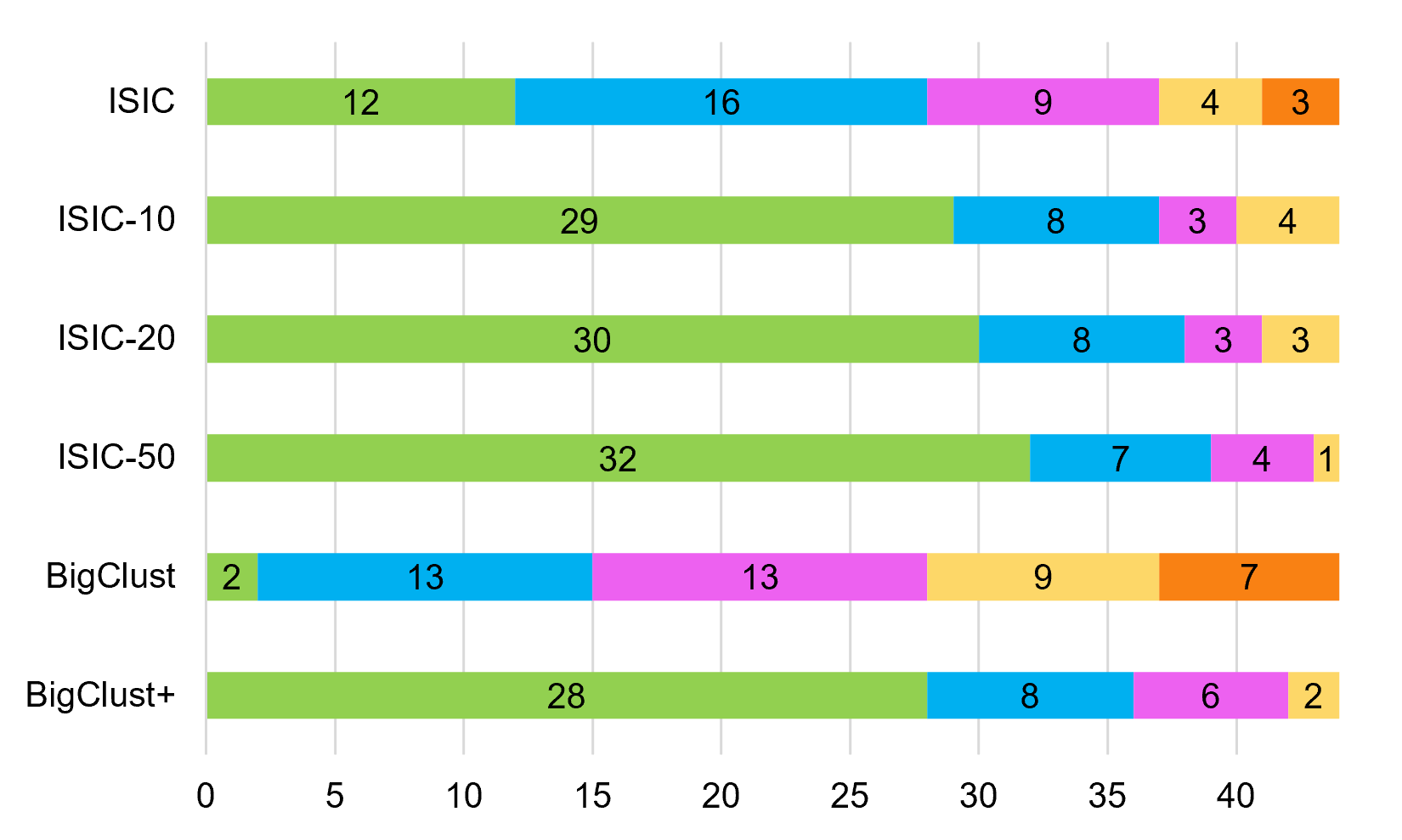}\label{fig:cpu_small}} 
   \subfigure[]{
   \includegraphics[width=0.48\textwidth]{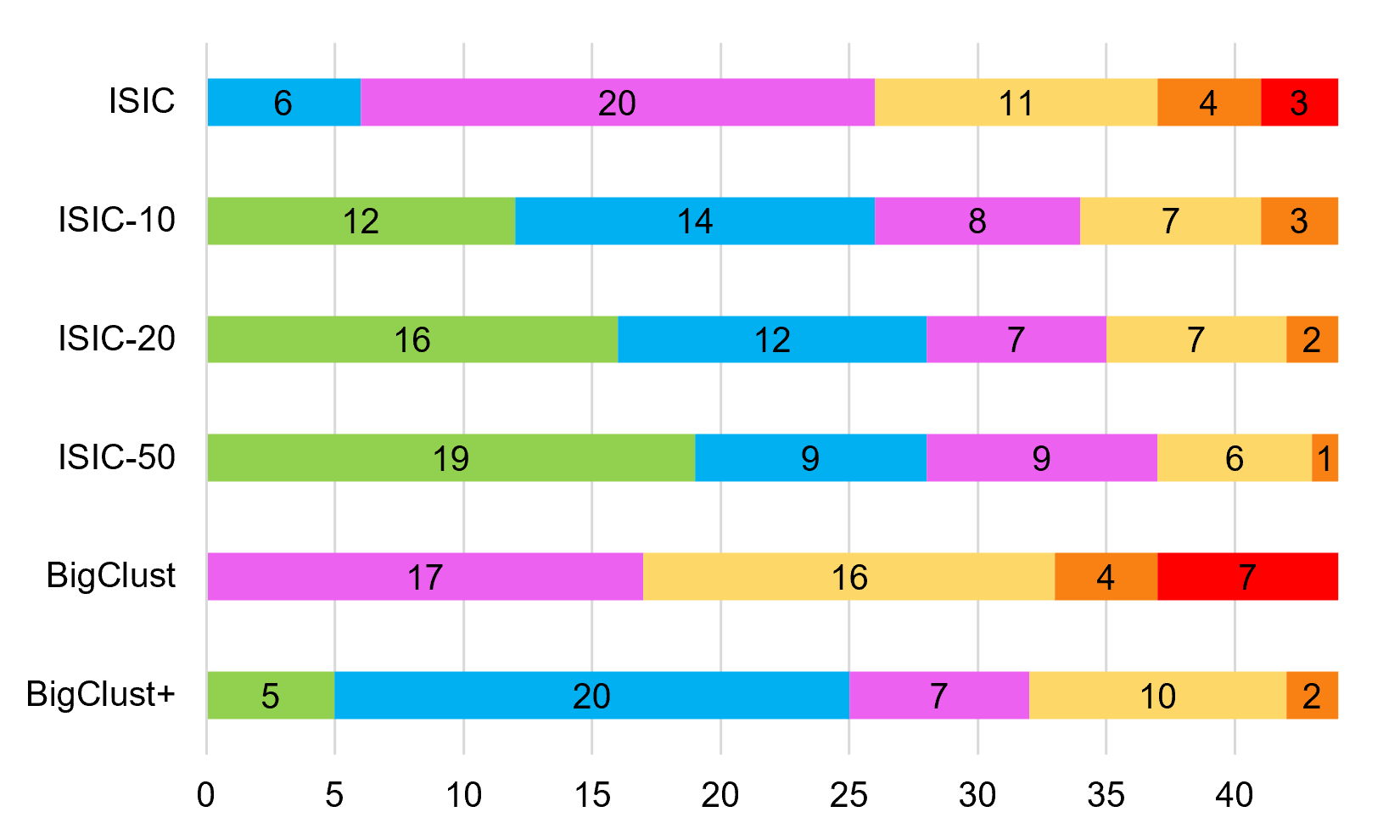}\label{fig:cpu_large}}
\caption{Cases categorized by CPU times for (a) $k$-clustering problems with $k=2,3,4,5$ and (b) $k$-clustering problems with $k=10,15,20,25$.}
\label{fig_cpu_subsample}
\end{figure}

Results of numerical experiments are summarized in Figures \ref{fig_summary1}-- \ref{fig_cpu_subsample}. In addition, the more detailed results are presented in Tables S5--S26 and Figures S1--S22 in Supplementary Material 2.

We use the following metrics to evaluate the clustering results:

\begin{enumerate}
\item \emph{Relative error.}
The cluster function value $f^k$ in \eqref{nonsmooth opt1} measures the average variation over all clusters. 
The relative error of an algorithm $\mathcal A$ is calculated as
    $$
     E_{\mathcal A} = \frac{\bar{f} - f_{\rm best}}{f_{\rm best}} \times 100\%,
    $$
where $\bar{f}$ is the cluster function value obtained by the algorithm $\mathcal A$ and $f_{\rm best}$ is the best-known cluster function value. Note that this relative error is identical to the corresponding relative error in the sum of squared errors (SSE).
For the comparison, we use the same best-known values $f_{\rm best}$ as in \cite{bigClust:2025}. Although smaller values may be known for some datasets at present, we keep the values from \cite{bigClust:2025} to make the results directly comparable.
For each dataset and each $k \in \mathcal{K}$, the minimum
($E_{\min}$), average ($E_{\rm aver}$), and maximum ($E_{\max}$)
relative errors are reported in Tables S5--S26 for the stochastic
algorithms, while a single relative error is reported for \lmbclustns.
The relative errors averaged over the values of $k \in \mathcal{K}$ (using
$E_{\rm aver}$ for the stochastic algorithms) are given in
Figures \ref{fig:sde} and \ref{fig:lde} for datasets with fewer
and more than 100\,000 samples, respectively.
%For each dataset (and each value of $k$ in case of \bigmeansns), ten runs are performed. The minimum ($E_{\min}$), average ($E_{\rm aver}$), and maximum ($E_{\max}$) relative errors are reported in Tables S5--S26 for $k \in \mathcal{K}$, while the average of $E_{\rm aver}$ (averaged over the values of $k$) is given in Figures \ref{fig:sde} and \ref{fig:lde} for less and more than 100\,000 samples, respectively.

\item \emph{Computational time.}
Computational times are given in seconds. Following \cite{bigClust:2025}, we distinguish between the time used for reading the data ($t_{\rm init}$), the time used for computing $k$ clusters ($t_k$), and the total computational time ($t_{\rm total}$). For \isicns, \sclustns, and \lmbclustns, the total time is calculated as $t_{\rm total}=t_{\rm init}+t_{25}$, whereas for \bigmeansns,
\[
t_{\rm total}=t_{\rm init}+\sum_{l \in \mathcal{K}} t_l,
\qquad
\mathcal{K}=\{2,3,4,5,10,15,20,25\}.
\]
Thus, $t_{\rm total}$ includes both data reading and clustering computations.
The times required to compute $k=10$ and $k=25$ clusters are summarized over the datasets in Figures~\ref{fig:sdt10} and \ref{fig:ldt10}, and Figures~\ref{fig:sdt25} and \ref{fig:ldt25}, respectively. Figure~\ref{fig_times} then illustrates in more detail the computational times $t_l$ ($l=1,\ldots,k$) of the incremental algorithms for different numbers of clusters on the Online News Popularity and Music Analysis datasets. Complete computational times for all datasets are provided in Tables~S5--S26 and Figures~S1--S22 in 
Supplementary Material 2.

\item \emph{Davies-Bouldin (DBI) and Dunn (DI) cluster validity indices.}
The DBI \cite{DavBou:1979} measures cluster compactness and separation, with smaller values indicating better clustering. The DI \cite{Dun:1974} measures the ratio between inter-cluster separation and intra-cluster spread, with larger values indicating better clustering. These indices can also help determine the appropriate number of clusters ($k$), with the minimum DBI and maximum DI indicating the preferred number of clusters. For more details see Supplementary Material 2. The DBI and DI are computed internally at each iteration of the incremental algorithms \isicns, \sclustns, and \lmbclustns. They are not reported for the non-incremental algorithm \bigmeansns. They are also not reported for the reduced-sample methods \isicns-\#, where $\#=10,20,50$, and \sclustns+, because these methods compute the DBI and DI using only part of the data, which may not provide accurate measures for the entire datasets. The results for the Protein Homology dataset are shown in Figure \ref{fig_dbi_and_di}, while the results for all datasets are presented in Figures S1--S22.
\end{enumerate}

Figures~\ref{fig:sde} and \ref{fig:lde} show that \lmbclust is the most accurate method, achieving the lowest average error in 14 out of the 22 datasets. However, it fails on two datasets, Cord-19 Embeddings and HEPMASS, due to memory issues. In the remaining eight datasets, \bigmeans achieves the lowest average errors. In particular, on the Gisette dataset, its average error is $-0{.}34$ (see Table~S6). These favorable results of \bigmeans should, however, be interpreted with some caution: in the datasets where \bigmeans performs best, the competing methods generally also achieve relatively small errors, whereas in six other datasets \bigmeans produces substantially larger errors than the competing methods, including very large errors in several cases and effective failures in two datasets, where the average relative error exceeds $10^6$ (see Figures \ref{fig:sde} and \ref{fig:lde}). Thus, \bigmeans can be considered the least consistent method in terms of accuracy.

The average errors of \isic and \sclust are very similar (see Figures \ref{fig:sde} and \ref{fig:lde}). The proposed \isic method achieves a smaller average error than \sclust in eight datasets, whereas \sclust performs better in the remaining 14 datasets. However, the differences between their average errors are generally small, with only a 0.23\% difference between their overall average errors.

Although \lmbclust can be considered the most accurate method, it is also the most time-consuming. When calculating 25 clusters, it is the slowest method in 16 out of 22 datasets (see Figures \ref{fig:sdt25} and \ref{fig:ldt25}) and it also fails in two datasets. In particular, for datasets containing more than 100\,000 data points, \lmbclust is the slowest method when calculating 25 clusters in all cases (see Figure \ref{fig:ldt25}). The main reason for the longer computational times of \lmbclust  compared with other methods is that \lmbclust processes the entire dataset, whereas the other methods use partial data.

There is also a clear pattern in how the computational times of the batch-based methods \isic and \sclust compare with those of \lmbclustns. The proportion of the batch size $m_\batch$ relative to the number of data points $m$ depends on the dataset size. When $m > 50\,000$, $m_\batch$ is always 2\% of $m$, whereas the proportion is higher for smaller datasets. For instance, in the ISOLET dataset, it is 13\%. Consequently, the computational time reduction achieved by \isic and \sclust compared with \lmbclust is smaller for datasets with fewer data points than for larger datasets. For example, when calculating 25 clusters in the D15112 dataset ($m=15\,112$), the computational times $t_{25}$ are 0{.}50 for \isicns, 3{.}91 for \sclustns, and 4{.}59 for \lmbclust (see Table S9). In contrast, for the Protein Homology dataset ($m=145\,751$), the corresponding times are 22{.}63 for \isicns, 50{.}30 for \sclustns, and 248.93 for \lmbclust (see Table S18). Thus, the larger the dataset, the greater the computational benefit of the batch-based approach.

A comparison with \bigmeans shows that, for datasets with $mn\le 50\,000\,000$ (see Table \ref{data}), \isic is more efficient in terms of computational time for all datasets when $k=10$ (see Figures~\ref{fig:sdt10} and~\ref{fig:ldt10}). In addition, for $k=25$, \isic is faster on most of these datasets, except on ISOLET, Covertype, and BitCoin (see Figures~\ref{fig:sdt25} and~\ref{fig:ldt25}). Here, the differences are rather small for Covertype and BitCoin, both of which have relatively few features. It is worth noting that datasets with $mn>50\,000\,000$ generally have a large number of features, which makes them more challenging for \isic (see Remark \ref{ordo}).

Furthermore, an important distinction is that, because \isic is an incremental algorithm, solving the $k$-clustering problem with $k=25$ also involves solving all intermediate problems with $k=2,\ldots,24$. In contrast, \bigmeans solves only the problem for the specified number of clusters. For example, for the Protein Homology dataset, \isic requires a total computational time of $t_{\rm total}=25{.}70$ to solve the 25-clustering problem and all intermediate $k$-clustering problems with $k=2,\ldots,24$. In contrast, solving the 25-clustering problem using \bigmeans requires a computational time of $t_{25}=40{.}16$, while solving the $k$-clustering problems with all $k \in \mathcal{K}$ (eight problems) separately requires $t_{\rm total}=324{.}54$.

Figures \ref{fig_summary1} and \ref{fig_summary2} show that \isic is faster than \sclust for calculating both 10 and 25 clusters across all datasets. The same pattern can be observed for all numbers of clusters in Figures \ref{fig_times} and S1--S22. For example, in the KEGG Metabolic dataset, the average computational times for calculating 25 clusters are 2{.}39 for \isic and 5{.}76 for \sclustns. Thus, \sclust requires 141\% more computational time than \isic (see Table S11). Although the absolute difference is less than three and half seconds, the relative increase is substantial. In the HEPMASS dataset, the difference is even more pronounced: $t_{25}$ is 673{.}78 for \isic and 3995{.}40 for \sclustns, corresponding to a 493\% increase (see Table S26). Overall, in terms of computational time, the proposed \isic method performs best, followed by \sclustns, \bigmeansns, and \lmbclustns.

Regarding the clustering validity indices DBI and DI, the values obtained by the incremental algorithms \isicns, \sclustns, and \lmbclust are very similar, particularly for smaller numbers of clusters (see Figures \ref{fig_dbi_and_di} and S1--S22). In particular, \cite{bigClust:2025} reports that \sclust and \lmbclust generally achieve similar or better validity indices than other incremental clustering algorithms, with smaller DBI values and larger DI values. This indicates that they are better able to identify well-separated clusters than the other algorithms. Thus, the proposed \isic algorithm also produces well-separated clusters and achieves solution quality comparable to that of the other incremental clustering algorithms. Furthermore, the correct number of clusters is not known for the Protein Homology dataset; however, $k=2$ is a natural choice if the two class labels used in the original classification task are interpreted as clusters. Our results also favor $k=2$, with $k=4$ emerging as the second-best alternative.

To sum up, \lmbclust achieves the highest accuracy in most datasets, but requires substantially more computational time than the other methods and fails on two datasets due to memory issues. The performance of \bigmeans is less consistent, with substantially larger errors on several datasets and generally longer computational times than \isic and \sclustns. Moreover, as \bigmeans is not incremental, evaluating multiple values of $k$ requires separate runs, further increasing its computational cost.

Despite using a less accurate starting-point selection procedure, \isic achieves accuracy comparable to \sclust while requiring considerably less computational time. This suggests that its optimization engine, \islmbmns, performs reliably despite the inexactness arising from stochastic computations. Overall, \isic provides a competitive alternative to existing MSSC methods for large-scale clustering, offering a favorable balance between solution quality and computational efficiency.

\paragraph{Results for reduced-sample methods}

Next, we analyze the reduced-sample methods \isicns-10, \isicns-20, \isicns-50, and \sclustns+ and compare them with their full-data counterparts, \isic and \sclustns. Since the reduced-sample approach is intended to improve computational efficiency for very large datasets, we restrict this analysis to the 11 datasets containing more than 100\,000 samples, for which the use of only a subset of the data can provide a meaningful computational advantage. The results are summarized in Figures~\ref{fig_error_subsample} and \ref{fig_cpu_subsample} (see Supplementary Material 2 for detailed comparisons), where the cases are categorized according to relative error and CPU time, respectively. Here, {\em a case} corresponds to one combination of a dataset and a prescribed number of clusters $k$. With 11 datasets and 8 values of $k$, each algorithm is evaluated on $11\times8=88$ cases. Further, the values of $k$ are divided into two groups: small ($k=2,3,4,5$) and large ($k=10,15,20,25$), giving 44 cases in each group.


Both \isic and \sclust achieve lower relative errors than their reduced-sample versions, but require longer CPU times (see Figures \ref{fig_error_subsample} and \ref{fig_cpu_subsample}). This is expected, as smaller sample and batch sizes reduce computational cost, potentially at the expense of clustering accuracy. Among the reduced-sample versions of \isicns, \isicns-10 achieves the lowest relative errors (see Figure \ref{fig_error_subsample}), with only slightly longer CPU times (see Figure \ref{fig_cpu_subsample}). Since the differences in computational time are small compared with those in relative errors, \isicns-10 offers the best trade-off between accuracy and computational cost among these variants.  %Detailed comparisons are provided in the supplementary material.

To further evaluate the performance of \isicns-10, we compare it with \sclustns+. In terms of CPU time, the two methods perform nearly identically: for small $k$, \isicns-10 solves 37 cases and \sclustns+ 36 cases in less than 5 seconds, while for large $k$, \isicns-10 solves 26 such cases, compared with 25 for \sclustns+. The main difference is therefore in accuracy. \isicns-10 achieves average relative errors below $0{.}05\%$ in 6 cases and above $50\%$ in only 7 cases, compared with 2 and 16 cases, respectively, for \sclustns+. Thus, \isicns-10 also outperforms \sclustns+ in accuracy while maintaining essentially the same computational cost.

Overall, \isicns-10 offers the best balance between clustering accuracy and computational efficiency among the reduced-sample variants, providing further computational savings with only a modest loss in accuracy compared with the full-data \isicns.

\section{Conclusions}\label{sec_conclusion}

In this paper, we introduced \isicns, an incremental clustering algorithm for solving MSSC problems in very large and big datasets. \isic further develops the incremental clustering framework through two new components: the greedy
starting-point algorithm \gspa and the stochastic inexact limited-memory bundle method \islmbmns. The first component, \gspans, replaces the computationally expensive multi-start procedure used in earlier incremental clustering approaches by directly
constructing a single promising starting center for each new cluster. Thus, it also eliminates the need to solve separate auxiliary clustering problems. The second component, \islmbmns, serves as the optimization engine throughout the incremental procedure. By building on \ilmbm \cite{inexactLMBM2026}, \islmbm explicitly accommodates inaccuracies in function values and subgradient information arising from stochastic computations.

The incremental structure of \isic produces solutions not only to the final $k$-clustering problem but also to all intermediate $l$-clustering problems, $l=1,\ldots,k-1$. An additional advantage of \isic is that it does not require full-data objective-function evaluations during the optimization process. The incremental procedure can operate on a sample redrawn at each incremental step, while \islmbm operates on randomly chosen batches within that sample. Consequently, unlike approaches that require repeated computations over the entire dataset, \isic can also be applied in settings where the complete dataset does not fit simultaneously into memory.

The numerical experiments demonstrate that \isic produces high-quality clustering solutions with substantially reduced computational effort. Its average errors are comparable to those of its closest predecessor, \sclustns, while its computational times are considerably lower. Compared with the deterministic \lmbclustns, \isic generally achieves somewhat lower accuracy, but requires substantially less computational time and has lower memory requirements. In comparison, \bigmeans produces substantially larger errors on several datasets and, in most cases, also requires more computational time than \isic and \sclustns. The fact that \isic achieves accuracy comparable to that of \sclust despite using a less accurate procedure for selecting the starting point suggests that its optimization engine, \islmbmns, is robust to the inexactness arising from stochastic computations. Overall, \isic provides a favorable balance between solution quality and computational efficiency, making it a competitive method for clustering large datasets. Furthermore, reduced-sample versions of \isic offer additional computational savings at the cost of a slight reduction in the accuracy of the solution.

\medskip
\noindent\textbf{CRediT authorship contribution statement}

\noindent
{\bf Napsu Karmitsa:} Conceptualization, Methodology, Software, Validation, Formal analysis, Investigation, Writing -- original draft.\\
{\bf Jenni Lampainen}: Conceptualization, Methodology, Investigation, Writing -- original draft, Visualization.\\
{\bf Kaisa Joki} and {\bf Marko M.\ Mäkelä}: Conceptualization, Methodology, Writing -- review and editing.\\
{\bf Antti Airola} and {\bf Tapio Pahikkala}: Conceptualization, Methodology, Formal analysis, Writing -- review and editing, Funding acquisition.

\medskip 
\noindent\textbf{Acknowledgement}
This work was financially supported by 
the University of Turku Graduate School (UTUGS) -- Doctoral Programme in Exact Sciences (EXACTUS), the University of Turku, 
the Research Council of Finland (Grant No.\ \#340140, \#340182, \#369549), 
and the European Union's Horizon Europe project Privacy Preserving Identity Management for Digital Wallet and Secure Data Sharing and Processing for Cyber Threat Intelligence Data (PRIVIDEMA, Grant Agreement No.\ 101167964).
Views and opinions expressed are however those of the authors only and do not necessarily reflect those of the European Union or the European Cybersecurity Competence Centre. Neither the European Union nor the European Cybersecurity Competence Centre can be held responsible for them.

\medskip
\noindent\textbf{Use of generative AI / AI-assisted technologies}
ChatGPT (GPT-5.6 Sol) was used solely for language polishing, primarily to improve grammar and readability, and not for generating scientific content. The authors reviewed and approved all edits and take full responsibility for the content of the manuscript.

\bibliographystyle{elsarticle-num}
\bibliography{references_PR_abbreviated}
%\bibliography{references}

%\section*{Appendix}\label{appendix}

\end{document}